\documentclass{article}

\usepackage{hyperref}
\usepackage{amsmath}
\usepackage{graphicx}
\usepackage{caption}
\usepackage{subcaption}
\usepackage{float}
\usepackage{siunitx}

\title{Olfactory Label Prediction on Aroma-Chemical Pairs}
\author{
Laura Sisson \\ \texttt{laura.h.sisson@gmail.com}
  \and 
  Aryan Amit Barsainyan \\ \texttt{aryan.barsainyan@gmail.com}
  \and
  Mrityunjay Sharma\textsuperscript{1,2} \\ \texttt{mrityunjay.csio24j@acsir.res.in} 
  \and
  Ritesh Kumar\textsuperscript{1,2}  \\ \texttt{riteshkr@csio.res.in}
}
\date{May 2024}
\newcommand{\desc}[1]{{\small``#1''}}
\begin{document}
\maketitle
\begin{center}
  \textsuperscript{1}
CSIR- Central Scientific Instruments Organisation, Sector 30-C,Chandigarh -160030, India  \\
  \textsuperscript{2}Academy of Scientific and Innovative Research (AcSIR), Ghaziabad- 201002, India \\
\end{center}
\section{Abstract}

The application of deep learning techniques on aroma-chemicals has resulted in models more accurate than human experts at predicting olfactory qualities. However, public research in this domain has been limited to predicting the qualities of single molecules, whereas in industry applications, perfumers and food scientists are often concerned with blends of many molecules. In this paper, we apply both existing and novel approaches to a dataset we gathered consisting of labeled pairs of molecules. We present graph neural network models capable of accurately predicting the odor qualities arising from blends of aroma-chemicals, with an analysis of how variations in architecture can lead to significant differences in predictive power.

\section{Introduction}
Carefully designed fragrances and flavors appear everywhere in our daily lives, like in our food, drinks and hygienic products. However, designing fragrant molecules is a laborious and time consuming iterative process. The forefront of quantitative olfactory research has been the hunt for new and explicable features used in the prediction of olfactory perception descriptors. The chemical space is huge (containing $\approx$ 10\textsuperscript{60} molecules) \cite{ruddigkeit2014expanding}, and ever-expanding.

Prior to the application of graph neural networks (GNNs) to odor prediction, researchers featurized aroma-chemicals based on specific molecular structures, like aromaticity and the presence of certain functional groups \cite{rossiter1996structure, keller2017predicting}. These approaches achieved decent success on benchmarks like the DREAM Olfactory Challenge \cite{keller2017predicting}; however, the adoption of GNNs to this domain led to significant improvements in the predictive power of contemporary models \cite{sanchez2019machine}. Instead of hand-engineered featurizations, recent methods for searching and analyzing the chemical space have utilized data-intensive stochastic or deep learning methods \cite{virshup2013stochastic, krems2019bayesian}. New machine learning architectures which operate on graphical or plain text representations of molecules have recently emerged \cite{jaeger2018mol2vec, coley2017convolutional, kearnes2016molecular, hull2001latent}, leading to improvements in computational understanding across drug discovery, material development, molecular property prediction, and de novo molecular design \cite{gomez2018automatic, wan2016deep, mikolov2013efficient, olivecrona2017molecular}. Recently, Lee et al. \cite{lee2023principal} used GNNs to predict aroma labels with high accuracy and precision, building a "perceptual odor map" from the underlying vector-embedding representations for each molecule. 

These breakthroughs in odor label prediction have enabled researchers to gain deeper insight into the relationship between scent and molecular structure. However, the non-linear relationships occurring in mixtures of aroma-chemicals have yet to be untangled. We adapt these deep learning methods and also apply new techniques to generate embeddings for blends of aroma-chemicals. Given the predominantly proprietary nature of research utilizing GNNs in chemistry, we needed to explore a variety of architectures to obtain these results.

This work represents a natural progression from the single molecule prediction task, where datasets of labeled molecules are individually processed, to a new blend-focused domain, where models are capable of predicting the odor of mixtures of molecules.

The contribution of this work is two fold: firstly, we present a carefully compiled odor mixture dataset wherein the perceptual descriptors of individual components and their blends are available; secondly, we present a set of publicly available GNN based algorithmic frameworks to predict the labels of the mixtures. We also present a number of selected experiments which showcase the efficacy of our approach, where our GNNs can transfer underlying models from blend prediction to single molecule prediction and vice versa \cite{tromelin2020exploring}.

\section{Methods}
\subsection{Data Collection}
To build the blended pair dataset, molecular structures (SMILES) and odorant labels were gathered from the GoodScents online chemical repository \cite{luebke2019good} to generate a dataset of aroma-chemical blends. While the GoodScents website cataloged \raisebox{-.9ex}{\textasciitilde}3.5k molecules, each aroma-chemical's page suggested a complementary odorant (called a "blender") that, when mixed together, yielded distinct aromas. Each molecule's page often contained more than 50 such recommendations, enabling us to gather over 160k molecule-pair data points

We built an adaptive web crawler that parsed the names, olfactory labels and recommended blenders for all odorants across GoodScents using the Python package BeautifulSoup. Non aroma-chemicals odorants were filtered out. In instances where aroma-chemicals lacked SMILES entries, or where the SMILES was unable to be parsed such data was excluded.
These malformed data points made up only 0.05\% of the total labeled pair dataset, and were therefore removed.

The collection of all molecule pairs in the database forms a meta-graph, where each node is itself a molecular graph, with edges between nodes if there are odor-labels for the blended pair of molecules. In order to ensure train/test separation, the meta-graph was carved into two components with the following requirements: firstly, in order to prevent distribution shift, each component must contain blended pair data points covering every label; secondly, in order to maximize the amount of usable data, the number of edges between the components (known as the edge-boundary degree) should be minimized, as these data points are thrown out to ensure train/test separation. Efficiently minimizing the edge-boundary degree is NP-hard in the general case \cite{seymour1994call}, although previous research demonstrates that some special cases can be solved in polynomial time \cite{kfoury2020efficient}.

\subsection{Dataset}
The dataset generated contains discrete labels for 109 olfactory notes\footnote{There was no data available for relative concentrations in the blends.}, which we standardized. Some entries contained pairs with no labels (marked as 'No odor group found for these') and these non-labeled pairs were removed. Additionally, \desc{anisic} was substituted with the more common \desc{anise}, \desc{medicinal,} (with a trailing comma) was adjusted to \desc{medicinal}, and \desc{corn chip} was replaced with \desc{corn}. These modifications led to a final set of 104 notes.

To examine the transfer learning capabilities of our models, we derived the single odorant datasets from Leffingwell \cite{leffingwell2005leffingwell} and GoodScents \cite{luebke2019good}, which were available as a combined dataset \cite{aryan_amit_barsainyan_ritesh_kumar_pinaki_saha_michael_schmuker_2023}.
\begin{figure}[H]
    \centering
    \begin{subfigure}[b]{0.30\textwidth}
        \centering
        \includegraphics[width=\textwidth]{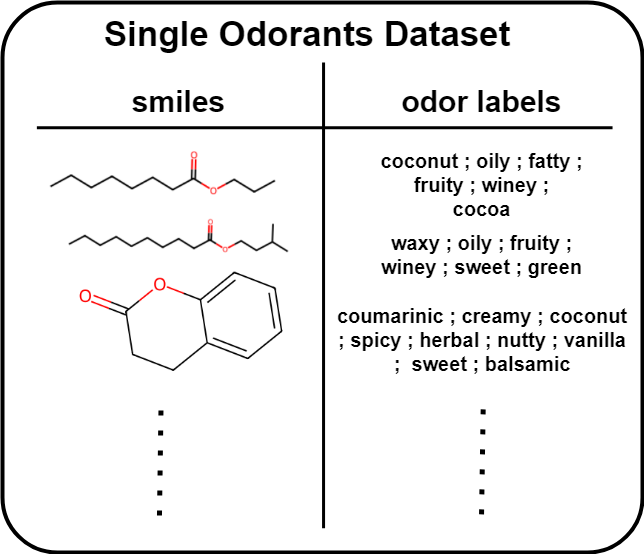}
        \caption{}
        \label{fig:figure1_subfig1}
    \end{subfigure}
    \begin{subfigure}[b]{0.30\textwidth}
        \centering
        \includegraphics[width=\textwidth]{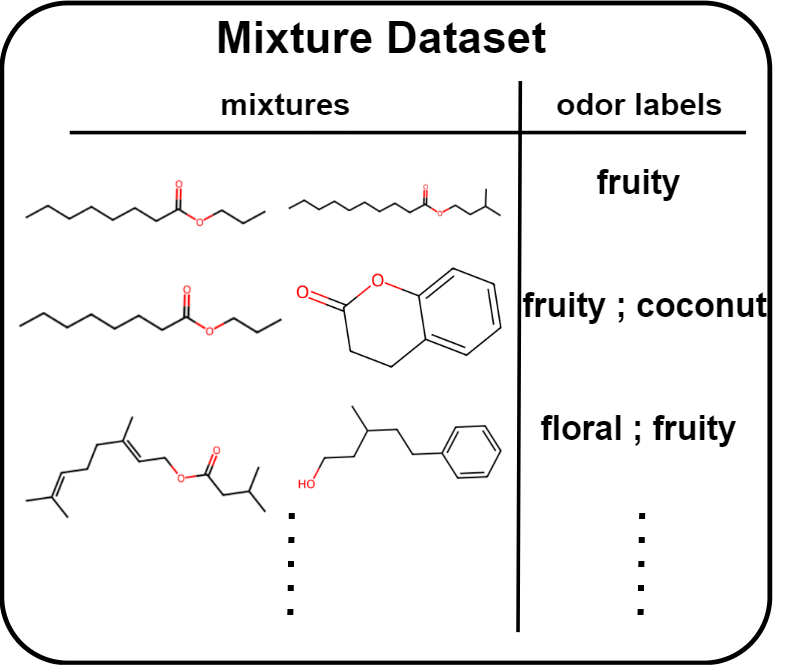}
        \caption{}
        \label{fig:figure1_subfig2}
    \end{subfigure}
    \begin{subfigure}[b]{0.30\textwidth}
        \centering
        \includegraphics[width=\textwidth]
        {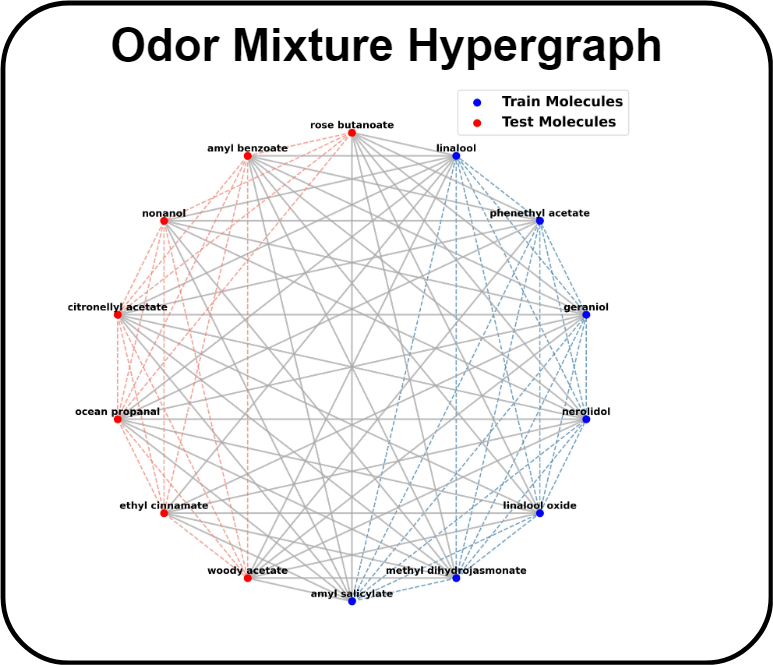}
       \caption{}
        \label{fig:figure1_subfig3}
    \end{subfigure} 
     \begin{subfigure}[b]{0.30\textwidth}
        \centering
        \includegraphics[width=\textwidth]{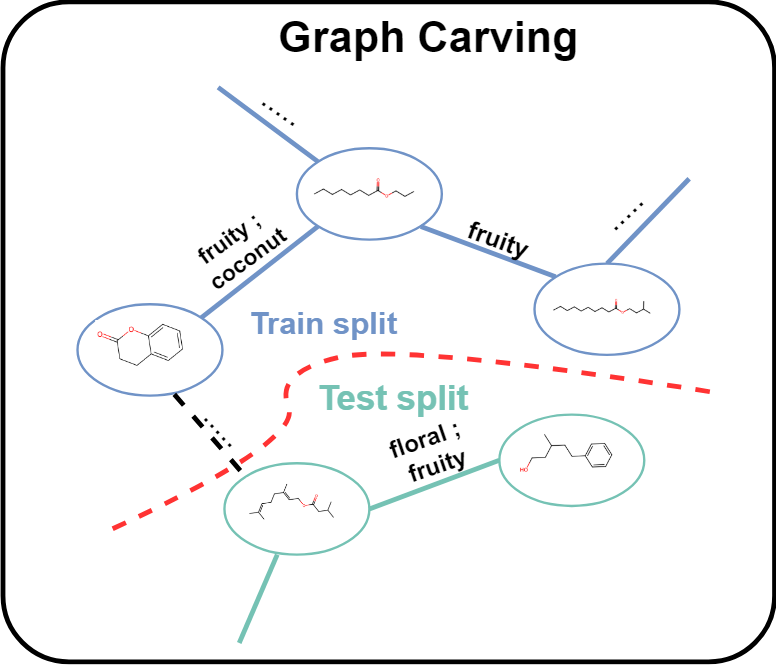}
        \caption{}
        \label{fig:figure1_subfig4}
    \end{subfigure}
     \begin{subfigure}[b]{0.30\textwidth}
        \centering
        \includegraphics[width=\textwidth]{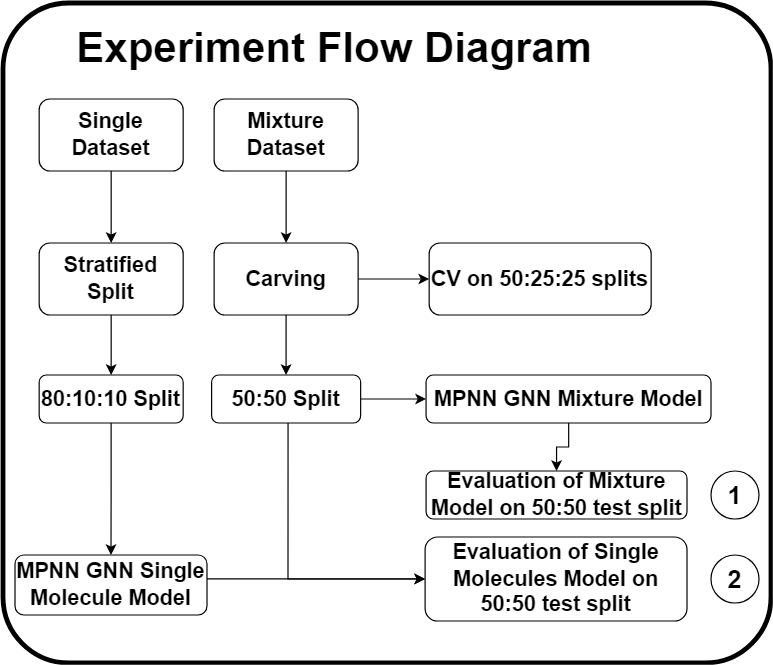}
       \caption{}
        \label{fig:figure1_subfig5}
    \end{subfigure}
     \begin{subfigure}[b]{0.30\textwidth}
        \centering
        \includegraphics[width=\textwidth]{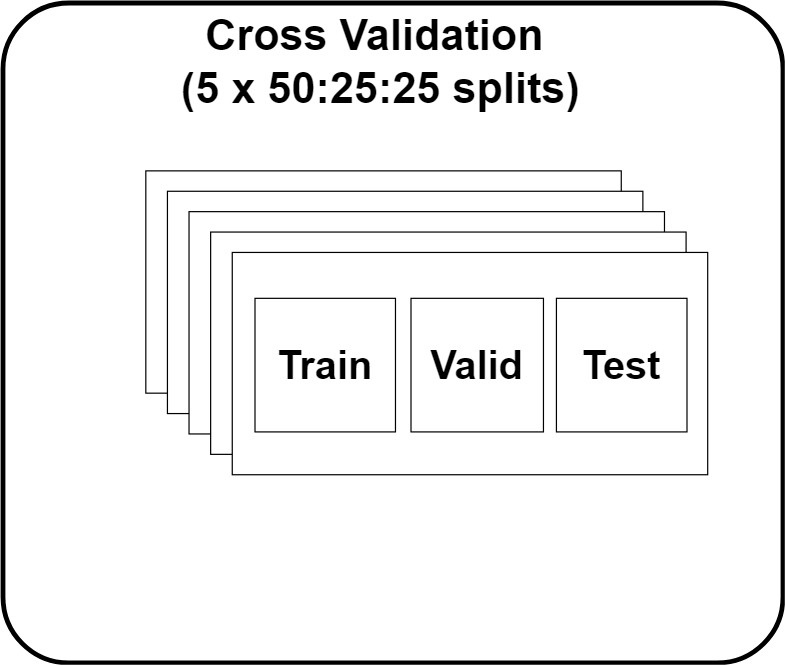}
       \caption{}
        \label{fig:figure1_subfig6}
    \end{subfigure}
     \begin{subfigure}[b]{0.30\textwidth}
        \centering
        \includegraphics[width=\textwidth]{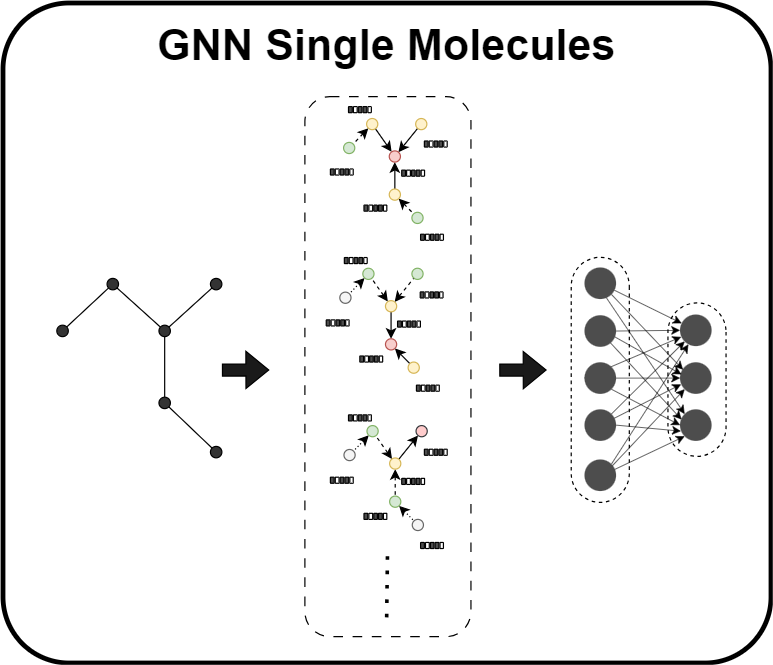}
       \caption{}
        \label{fig:figure1_subfig7}
    \end{subfigure}
     \begin{subfigure}[b]{0.30\textwidth}
        \centering
        \includegraphics[width=\textwidth]{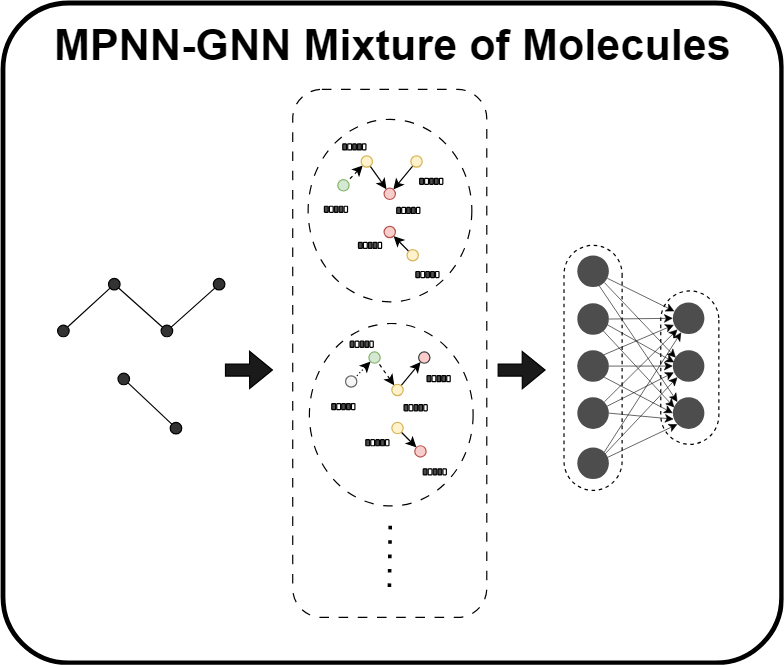}
       \caption{}
        \label{fig:figure1_subfig8}
    \end{subfigure}  
    \begin{subfigure}[b]{0.30\textwidth}
        \centering
        \includegraphics[width=\textwidth]{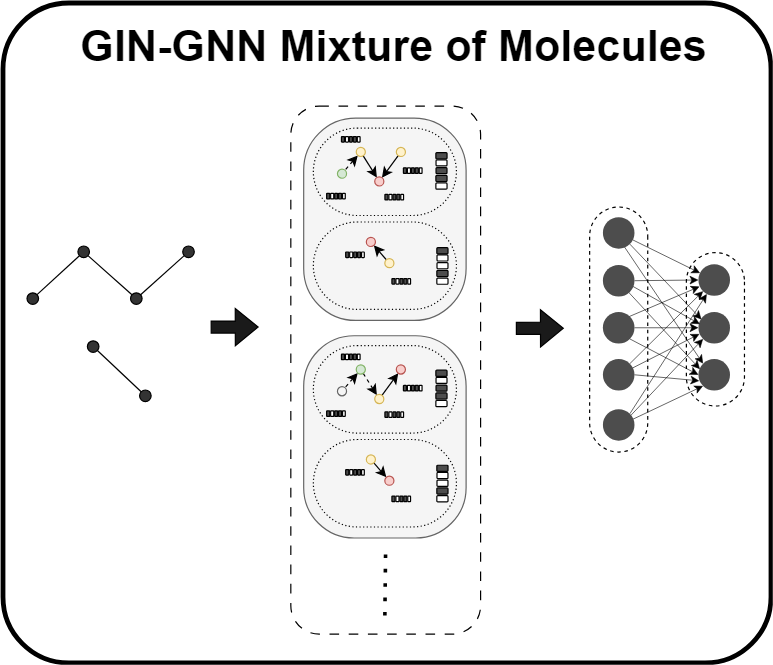}
       \caption{}
        \label{fig:figure1_subfig9}
    \end{subfigure}  
    \caption{\small\textbf{Methodology}\\ 
    \textbf{(a and b) Non-linear relationship between the qualities of constituent aroma-chemicals and the overall blend.} Although the same molecules appear across the single and mixture datasets, emergent notes appear when molecules are combine, and other notes become muted in the blend.\\
    \textbf{(c) Sample of the densest region of the blended pair meta-graph.} Here, 0.5\% of the total meta-graph nodes, consisting of 7 train molecules (in blue) and 7 test molecules (in red) are visualized, with an average degree of 6. Because there are many data points/edges per molecule, the meta-graph is dense and thus difficult to carve.\\
    \textbf{(d) Graph carving schematic.} The carving algorithm aimed to maximize the number of usable pairs without causing distributional shift in labels.\\
    \textbf{(e \& f) Schema of experiment.} The entire \textbf{(e) optimization and training pipeline} used in this paper, including \textbf{(f) 5-fold splits} of 50:25:25 train/test/validate splits used for hyper-parameter optimization.\\
    \textbf{(g) GNN predictions on single molecules.} Message passing layers are applied across the molecular graph, and then followed by a readout phase and a multi-layer perceptron (MLP) in order to predict the final label.\\
    \textbf{(h) MPNN-GNN predictions on blended pairs}, the molecular graphs are treated as one graph, with a combined readout and MLP as above.\\
    \textbf{(i) GIN-GNN predictions on blended pairs.} The molecule graphs have separate message passing and readout steps, and are combined only before the MLP. 
    } 
    \label{fig:figure1}
\end{figure}

\subsection{Graph Carving}
We carved the meta-graph by randomly partitioning the set of molecules into train and test splits. The carving algorithm was repeated until we generated a carving with at least one train and test data point for every label. With a 50:50 split, this took 90k iterations. We experimented scoring potential graph carvings using the Kullback–Leibler divergence between the train/test odor-label distribution and the distribution across the entire graph. However, we decided to optimize the number of usable data points rather than the similarity between components. Raising the fraction of training data meant that less data points were discarded, but also meant that it became impossible to carve a test set of molecules which covered all labels. Therefore, the 50/50 split was selected. This resulted in a final dataset with 44k training pairs and 40k test pairs, discarding 83k data points to satisfy the separation requirements. Out of the 109 odor labels, only 74 appeared across enough molecules to be carved. We attempted 250k further carvings, but never one that covered more labels.

\subsection{Cross Validation}
In order to test and identify the best performing models, we performed various sets of experiments on the carved train and test components of the datasets. These experimental procedures are documented in Fig. \ref{fig:figure1}.

\begin{enumerate}
  \item \textbf{5-fold split and hyperparameter search:} We conducted further searches to obtain five train-test-split carvings, which were fully separated as above. Because each component had less constituent molecules, the number of odor labels that were possible to carve was only the range of 41 to 44 per fold. We used these 5-fold splits to conduct hyperparameter optimization, hypothesizing that the best models would transfer well to the original 74 label 50:50 carving.
  \item \textbf{50:50 split on the model based on mixture data:} In this experiment we trained 3 MPNN-GNN models, with different random seeds, on the original 50:50 train-test split using the best hyperparameters found. 
\end{enumerate}

\section{Architecture}
\subsection{Models}
We trained a variety of GNN models for predicting the blended odor labels from the structures of pairs of aroma-chemicals. These models were derived from two primary architectures.

Firstly, we developed a novel Graph Isomorphism Network (GIN) \cite{xu2018powerful} based model, which independently generated embeddings for each molecule in a pair. These embeddings were combined during a final blended-pair prediction step. For detailed analysis of the GIN-GNN model, see section 1.2.1 of the supplementary information.

Secondly, we trained several variations on the Message Passing Neural Network (MPNN) inspired by Lee et al \cite{lee2023principal}. Here, the molecular structures for pairs of molecules were grouped into a single graph before being fed into the message passing layers. For the detailed analysis of the MPNN-GNN model, refer section 1.2.2 of the supplementary information.

\section{Results}
\subsection{Blended Pair Prediction}
To measure the predicting powers of the various models, we used the area under the receiver operating characteristic curve (AUROC) for each odor label. To compare the results, we took the micro-average across all test data points. The MPNN-GNN model achieved a mean AUROC of 0.77, with the GIN-GNN scoring a slightly lower mean AUROC of 0.76. For context, a naive 0-R model using the mean frequency of each label across all molecules as a constant prediction achieves an AUROC of 0.5 for every label, by definition. As a baseline model, we generated 2048-bit Morgan fingerprints (MFP) with $radius = 4$ for each molecule in a pair, which were concatenated and then used as input to a logistic regression to predict the blended pair's labels. 

While the GIN-GNN model predicts some labels very accurately, it significantly underperforms compared to the random baseline on other labels. On the other hand, the MPNN-GNN model performs consistently across all labels. One of the easiest descriptors to predict was \desc{alliaceous} (garlic), reflecting previous work \cite{lee2023principal} which suggested that this note straightforwardly correlated with the presence of sulfur in the molecule. Unlike in previous work, our models accurately predicted the label \desc{musk}, which is normally difficult to predict as it occurs across many different structural classes of molecules. However, direct comparison between benchmarks is not straightforward, as previous work predicted continuous ratings for odor, and our dataset contains discrete labels. There were some descriptors (like \desc{orris} and \desc{earthy}) which none of the molecules were able to accurately predict.
 
\subsection{Single Molecule Prediction}
We also measured the performance of our models on the single molecule prediction task. To adapt the GIN-GNN model to this task, we generated graph-level embeddings for each molecule and trained a logistic regression classifier to predict the same 74 odor labels. Because the graph-level and original pair-level embeddings were of different dimensionalities,the MLP portion of the architecture could not be transferred. For the MPNN-GNN, the only modification necessary was inputting a single molecule instead of a pair of molecules into the message passing phase. The entire trained architecture was able to be reused. On the single molecule task, the MPNN-GNN achieved a mean AUROC score of 0.89, while the GIN-GNN and Morgan Fingerprint models achieved scores of 0.85 and 0.82 respectively.

The significant improvement of all models on the single molecule prediction task, as compared to the blended pair task suggests that the former task is much harder than the latter. We hypothesize the widened gap between the MPNN-GNN and the GIN-GNN on this task is due to the fact that the GIN-GNN's prediction layers could not be reused.

Regardless, the descriptor \desc{alliaceous} remained easy to predict across the board, but surprisingly, \desc{musk} was one of the easiest descriptors to predict. In our dataset, \desc{musk} molecules, regardless of structural class, were often combined with each other in blends. This likely produced similar embeddings in our GNN models for the molecules, even though they were structurally dissimilar. This provided an advantage over previous work, where models had to learn the olfactory similarity for \desc{musk} molecules from their labels alone. One notable exception was the label \desc{aromatic}, which the GIN-GNN and Morgan Fingerprint models failed to accurately predict. 

The best hyperparameter values derived from experiments conducted on the combined Leffingwell-GoodScents dataset for both the blended pair and the single molecule trained model tasks are detailed in Tables S2 and Table S3 in the supplementary information.

\begin{figure}[H]
    \centering
    \begin{subfigure}[b]{0.9\textwidth}
        \centering
        \includegraphics[width=\textwidth]{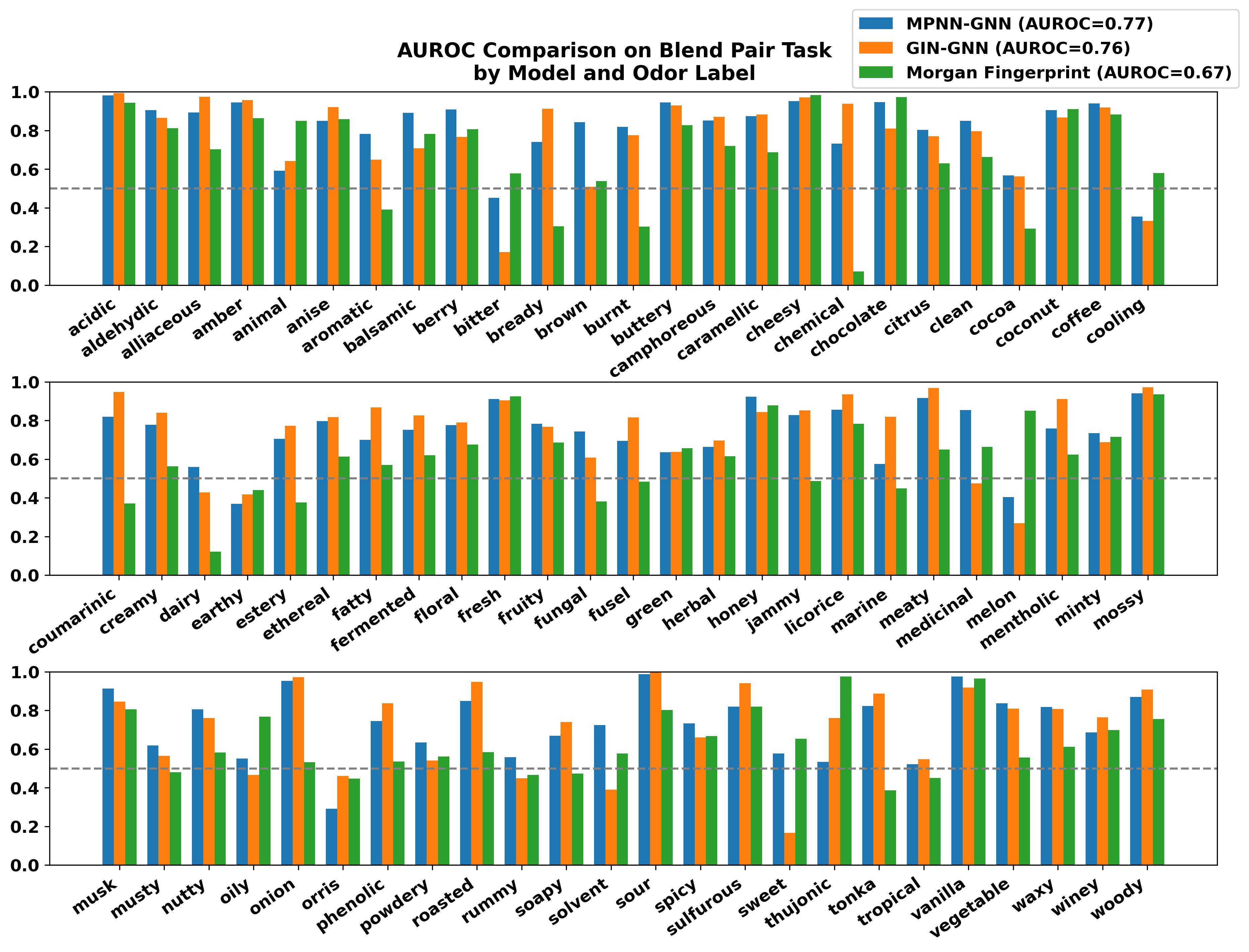}
        \caption{}
        \label{fig:molpairs1}
    \end{subfigure}
    \hfill
    \begin{subfigure}[b]{0.9\textwidth}
        \centering
        \includegraphics[width=\textwidth]{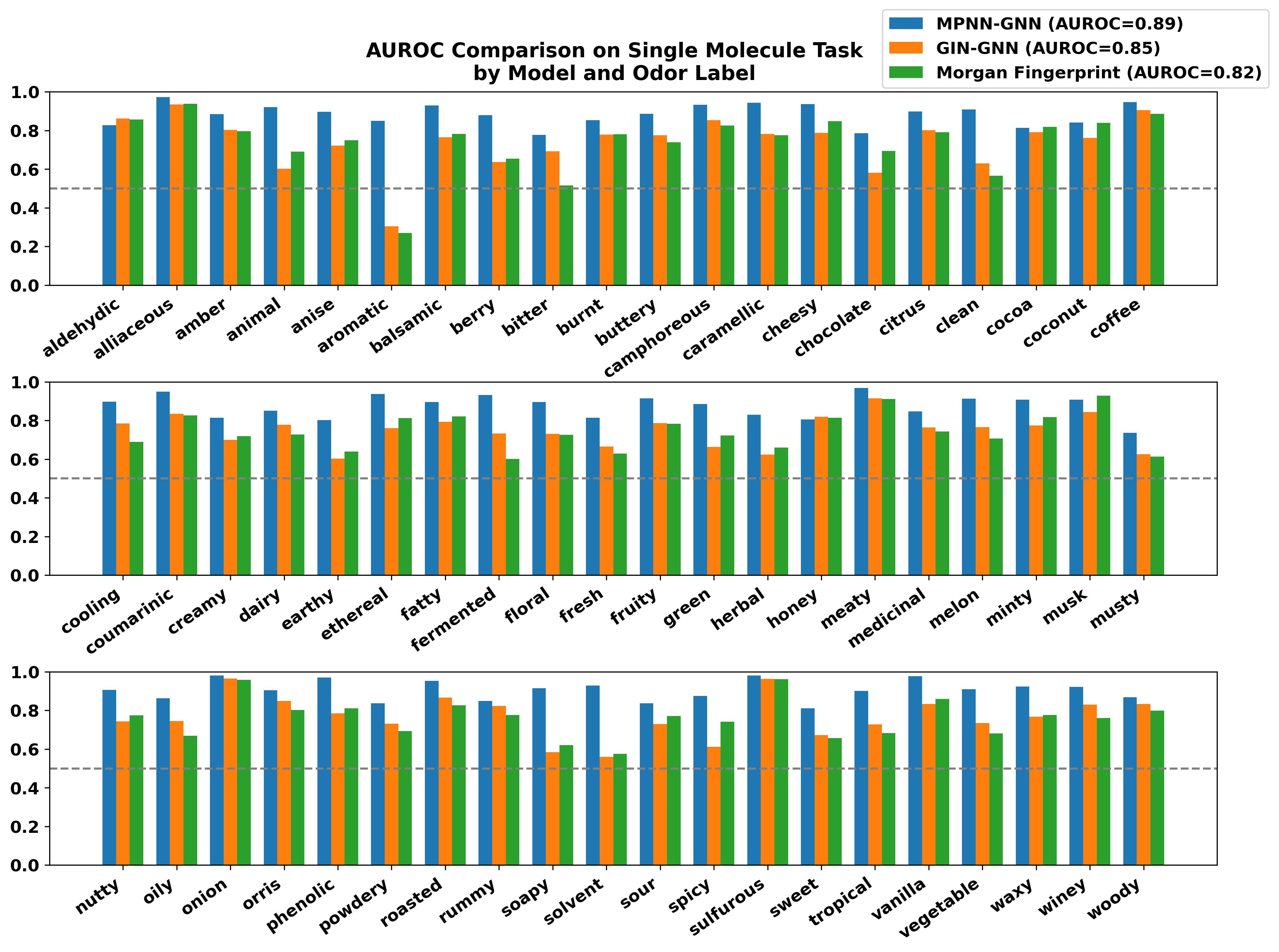}
        \caption{}
        \label{fig:molpairs2}
    \end{subfigure}
    \caption{\small\textbf{Predictive power of our GNN models and the Morgan Fingerprint baseline across all labels with random baseline (dashed line).}\\
    \textbf{(a) Blended pair task AUROC scores}, by descriptor.\\
    \textbf{(b) Single molecule task AUROC scores}, by descriptor.
    }
    \label{fig:combined}
\end{figure}

\section{Discussion}
\subsection{MPNN-GNN Performance}
\begin{figure}[H]
 
    \centering
    \begin{subfigure}[b]{0.30\textwidth}
        \centering
        \includegraphics[width=\textwidth]{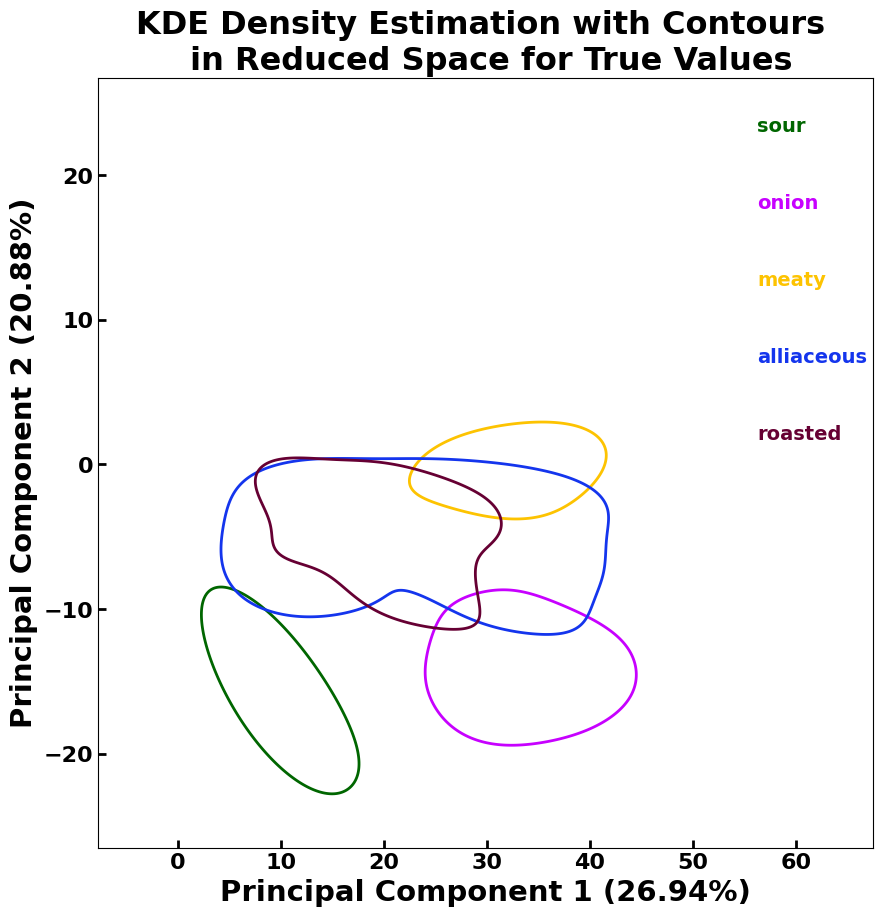}
        
        \caption{}
        \label{fig:Kde_Plot_subfig1}
    \end{subfigure}
    \begin{subfigure}[b]{0.30\textwidth}
        \centering
        \includegraphics[width=\textwidth]{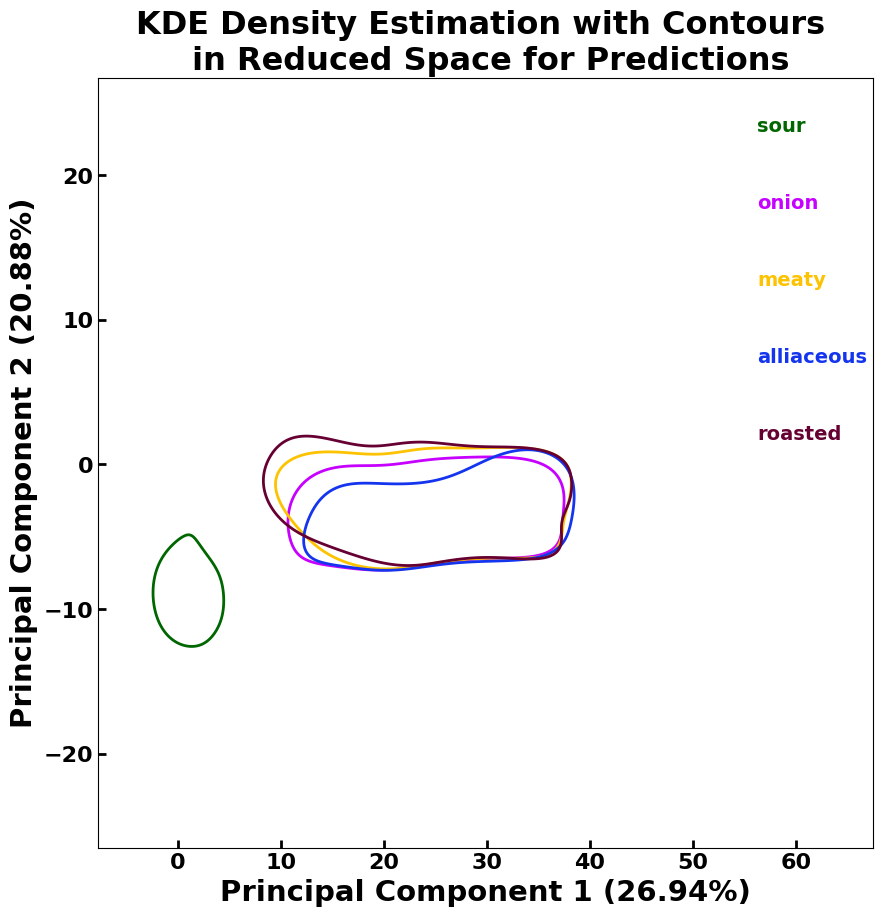}
        
        \caption{}
        \label{fig:Kde_Plot_subfig2}
    \end{subfigure}
    \begin{subfigure}[b]{0.30\textwidth}
        \centering
        \includegraphics[width=\textwidth]
        {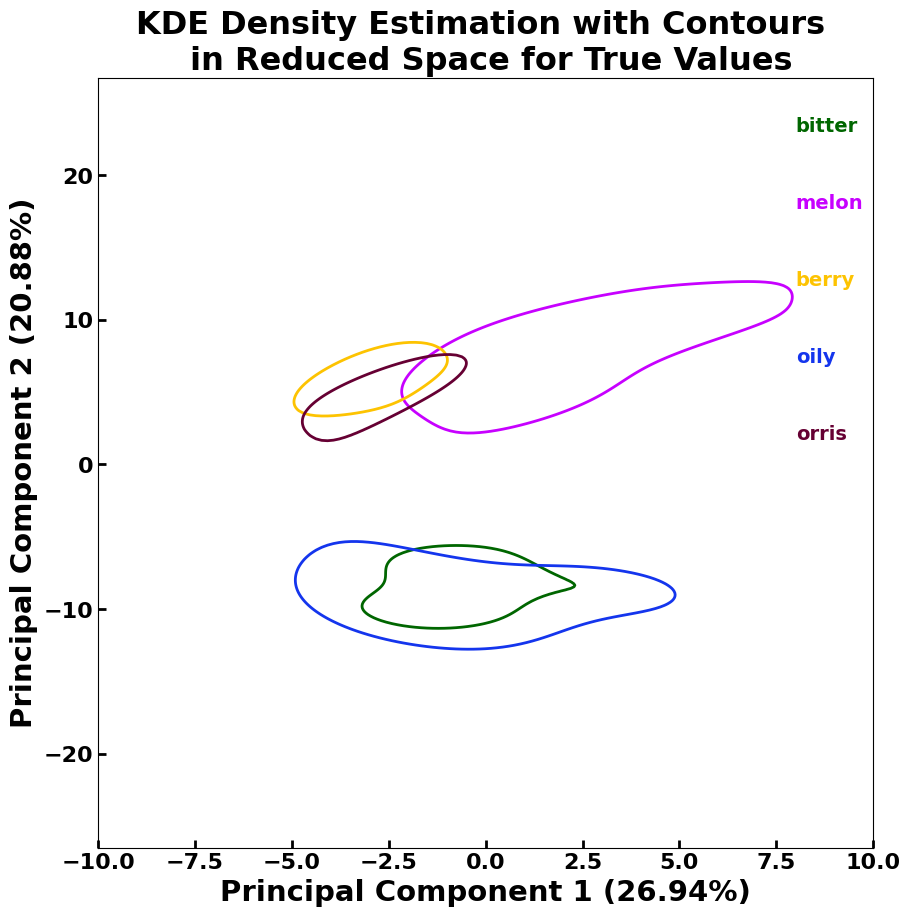}
       \caption{}
        \label{fig:Kde_Plot_subfig3}
    \end{subfigure} 
     \begin{subfigure}[b]{0.30\textwidth}
        \centering
        \includegraphics[width=\textwidth]
        {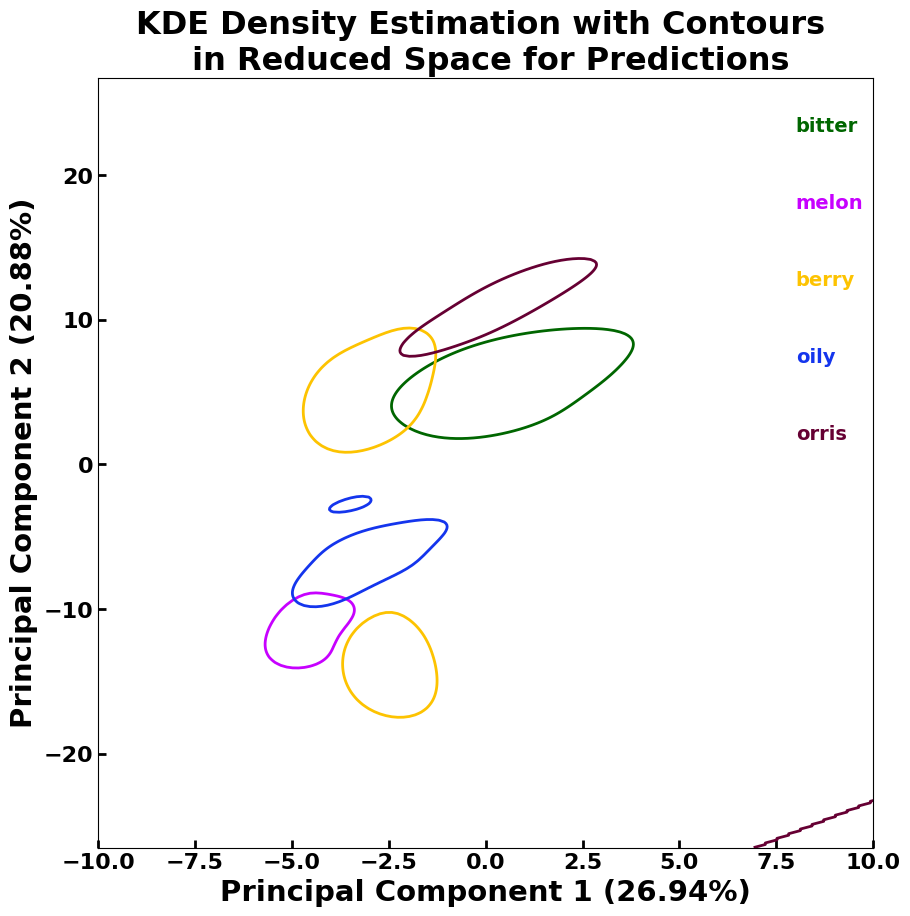}
       \caption{}
        \label{fig:Kde_Plot_subfig4}
    \end{subfigure} 
      \begin{subfigure}[b]{0.30\textwidth}
        \centering
        \includegraphics[width=\textwidth]{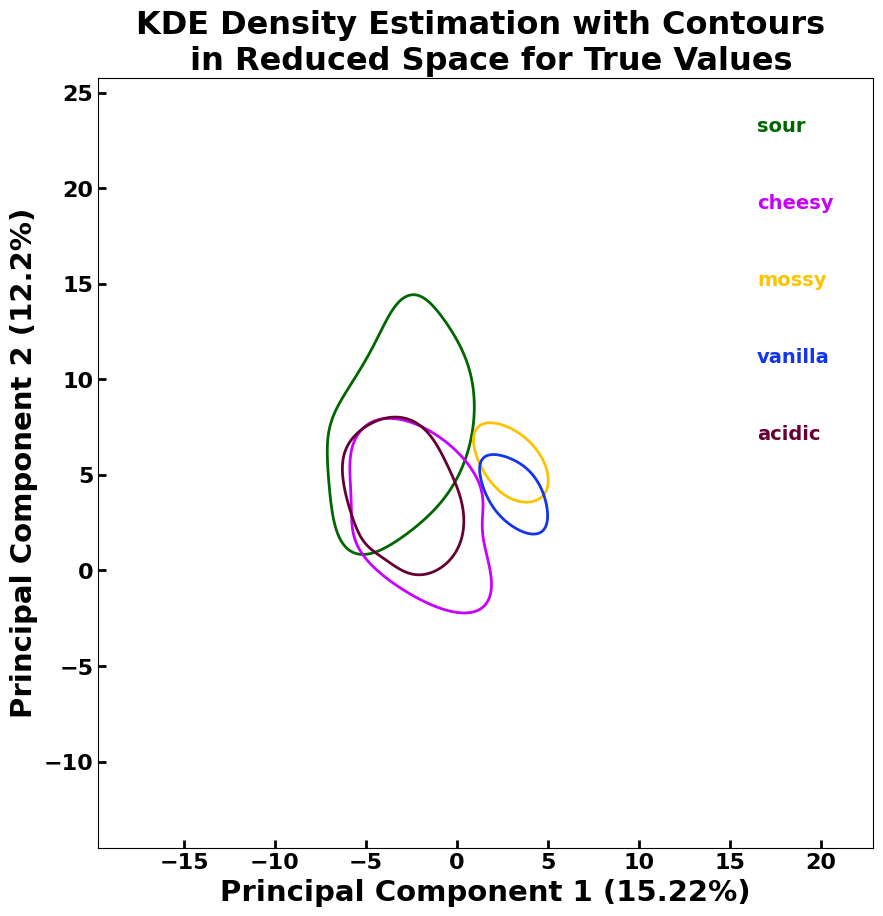}
       \caption{}
        \label{fig:Kde_Plot_subfig5}
    \end{subfigure}
     \begin{subfigure}[b]{0.30\textwidth}
        \centering
        \includegraphics[width=\textwidth]{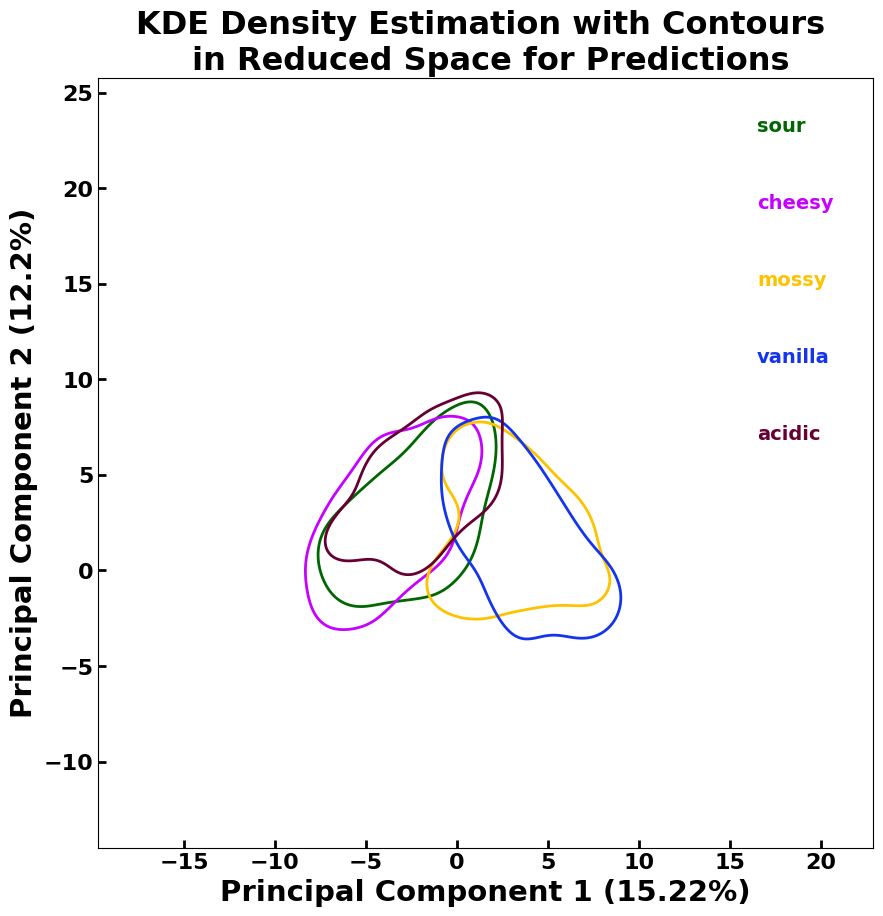}
       \caption{}
        \label{fig:Kde_Plot_subfig6}
    \end{subfigure}
     \begin{subfigure}[b]{0.30\textwidth}
        \centering
        \includegraphics[width=\textwidth]{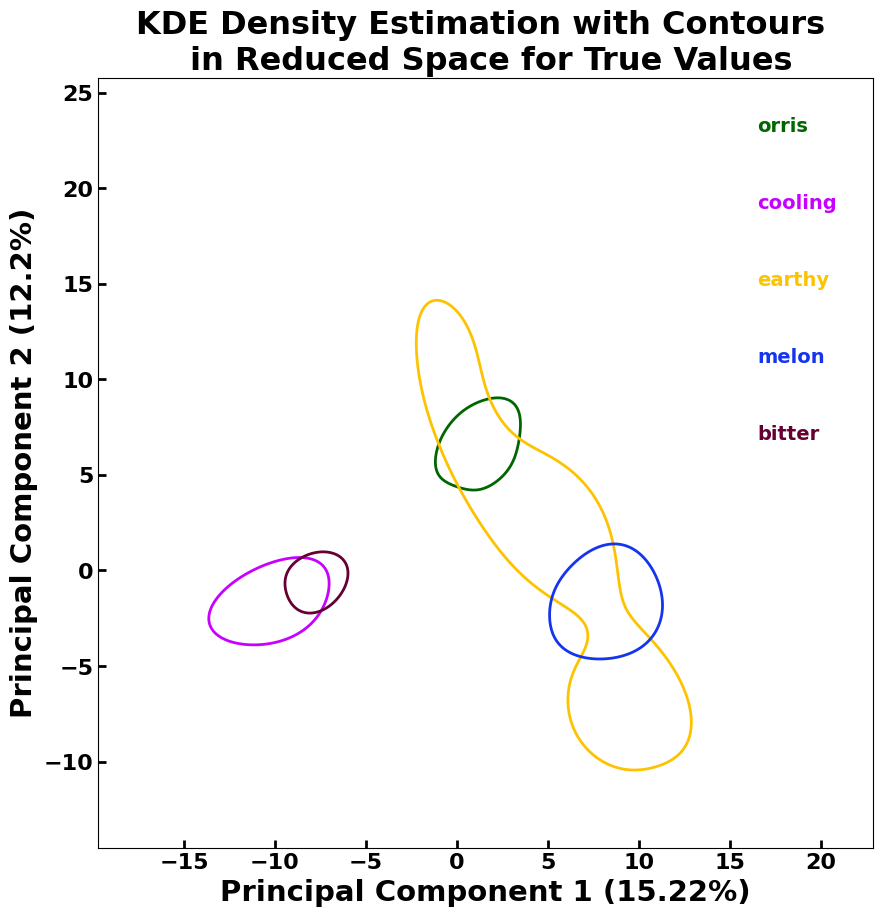}
       \caption{}
        \label{fig:Kde_Plot_subfig7}
    \end{subfigure}
     \begin{subfigure}[b]{0.30\textwidth}
        \centering
        \includegraphics[width=\textwidth]{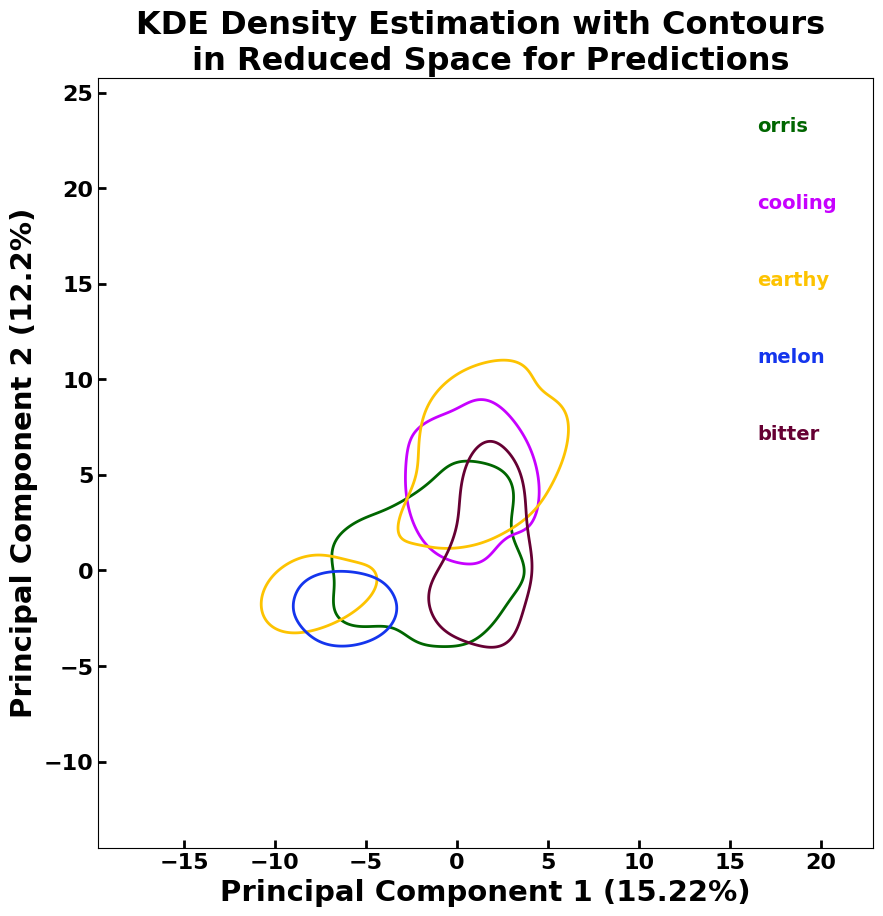}
       \caption{}
        \label{fig:Kde_Plot_subfig8}
    \end{subfigure}
    \caption{\small\textbf{Analysis of Odour labels by conducting experiments.}\\
    \textbf{(a) KDE plots for top 5 descriptors} by predictive accuracy in the training set of single molecule task.\\
    \textbf{(b) KDE top 5 single molecule as above}, for test set.\\
    \textbf{(c) KDE plots for bottom 5 descriptors} by predictive accuracy in the training set of single molecule task.\\
    \textbf{(d) KDE bottom 5 single molecule as above}, for test set.\\
    \textbf{(e) KDE plots for top 5 descriptors} by predictive accuracy in the training set of blended pair prediction task.\\
    \textbf{(f) KDE top 5 blended pair as above}, for test set.\\
    \textbf{(g) KDE plots for bottom 5 descriptors} by predictive accuracy in the training set of blended pair task.\\ \textbf{(h) KDE bottom 5 blended pair as above}, for test set.
    }
    \label{fig:Kde_Plot}
   
\end{figure}

Several experiments were conducted to understand the MPNN-GNN's performance on individual labels, focusing on the best and worst performing odor labels. Our goal was to analyze how model predictions correlated with the co-occurrence patterns of odor descriptors in blended pairs across both the training and test sets. Figure \ref{fig:Kde_Plot} shows the kernel density estimation (KDE) plots of the MPNN-GNN's most accurate and least accurate predictions on both the single molecule and blended pair prediction task, alongside the ground-truth values. 

Interestingly, the MPPN-GNN's predictive scores for the top 5 and bottom 5 are not correlated with the descriptors' occurrences in the training data, contradicting the findings of Lee et al. \cite{lee2023principal}. We conducted detailed analysis for a small set of descriptors:
\begin{itemize}
\item \textbf{\desc{orris}}: Though \desc{orris} has significant representation in the training and test sets, an imbalance in the data favors \desc{spicy} (both individually and in co-occurrence with \desc{orris}), potentially causing the model to prefer \desc{spicy} as the primary descriptor for mixtures containing \desc{orris}.
\item \textbf{\desc{cooling}}: In the test set, 75\% of the \desc{cooling} target distribution is also labeled \desc{green}, but the model predicts 50\% as \desc{estery} and 30\% as \desc{fruity}.
\item \textbf{\desc{green}}: The MPPN-GNN predicts \desc{green} as occurring in less than 2\% of test molecules, despite the label's major presence in the training and test sets. This could again be attributed to the individual co-occurrence of \desc{green} label with \desc{fruity} and \desc{estery} in the overall database.
\item \textbf{\desc{earthy}}: Similarly for \desc{earthy} labels, the model predicts 40\% as \desc{fatty} and 30\% as \desc{chocolate}, despite many \desc{earthy} pairs occurring in the training and test sets. \desc{Waxy} is well-predicted for mixtures smelling both \desc{earthy} and \desc{waxy}. 
\end{itemize}
There seems to be semantic spillover across labels with regards to their co-occurrences in both the blended pair and single molecule tasks. To a certain degree, this is expected, as olfactory labels are often nebulous and overlapping. To remedy this, further research into a definitive olfactory vocabulary must be conducted, where all descriptors are semantically distinct from each other \cite{lee2023principal,kumar2015understanding}. For further detailed analysis of the performance across labels, see "MPNN-GNN Performance" in the supplementary information.
 
\subsection{GIN-GNN Performance}
Across both the blended pair and single molecule prediction task, the GIN-GNN underperforms compared to the MPNN-GNN. We hypothesize this is primarily due to a number of differences in the architectures.

Firstly, molecules are fed into the GNNs in different ways. As stated above, the MPNN-GNN applies the message passing function across the pairs of molecules as if they were a single graph, with two disconnected components. The readout layer combines all the nodes across both graphs simultaneously. From there, additional feedforward layers generate the prediction, and this feedforward neural network can be reused for the single molecule prediction task. On the other hand, the GIN-GNN generates component embeddings separately and concatenates them together prior to the final feedforward layers. We hypothesize that this leads to weaker underlying representations for component molecules in the GIN-GNN, as shown in the performance comparison in Figure \ref{fig:molpairs1} and \ref{fig:molpairs2}, the difference is especially noticeable for the single molecule prediction task, where a new feedforward network must be trained from scratch. 

We also propose that the additional information provided by edge-conditioning in the MPNN-GNN results in a higher predictive capacity than in the GIN-GNN, which updates the hidden states of nodes based only on the node's own features.

Finally, we note that the set2set readout method provided better results than simply concatenating the mean and max pooling results of the node level embeddings.

\subsection{Embedding Space}
Although the relationship between the odors of individual molecules and the blended pairs in which they occur is non-linear, previous work has shown that embedding spaces can represent non-linear relationships with linear transformations. 

In order to explore the relationships between the latent space of individual molecules and that of the blended pairs, we generated three vector embeddings per data point: $\mathbf{e}_1$ (the embedding of the first molecule in the pair), $\mathbf{e}_2$ (the embedding of the second molecule in the pair), and $\mathbf{e}_p$ (the embedding of blended pair), using the MPNN-GNN. To determine if the blended pair embeddings were simply linear combinations of the constituent embeddings, we fit a number of linear regression models, one per pair in the dataset:
\[
\alpha_1 \cdot \mathbf{e}_1 + \alpha_2 \cdot \mathbf{e}_2 = \mathbf{e}_p
\]
where $\alpha_1$ and $\alpha_2$ are constant coefficients. We measured how well the linear regression model fit the relationship using the coefficient of determination ($r^2$). Across all blended pair data points, the average ($r^2$) was 0.47. While there was a significant portion of the relationship between constituent and blended pair embeddings that could be explained linearly, the majority was more complex.
Notably, $\alpha_1$  and $\alpha_2$  were inversely correlated, meaning that while a significant portion of blended pair embeddings were equal combinations of each constituent molecule, as the influence of one molecule increased in the blend, the other molecule’s influence decreased. This may be explained by the phenomena of additive or subtractive blending, wherein some molecules intensify the certain notes when combined, while other molecules may mute each other in combination.

The process was repeated with the GIN-GNN, though the component embedding space (D=100) and the pair embedding space (D=77) could not be directly compared, so we reduced the component embedding space to the same dimension as the pair embedding space, and fit linear regressors as above.

\begin{figure}[ht]
    \centering
    \begin{subfigure}[b]{.45\textwidth}
        \centering
        \includegraphics[width=\textwidth]{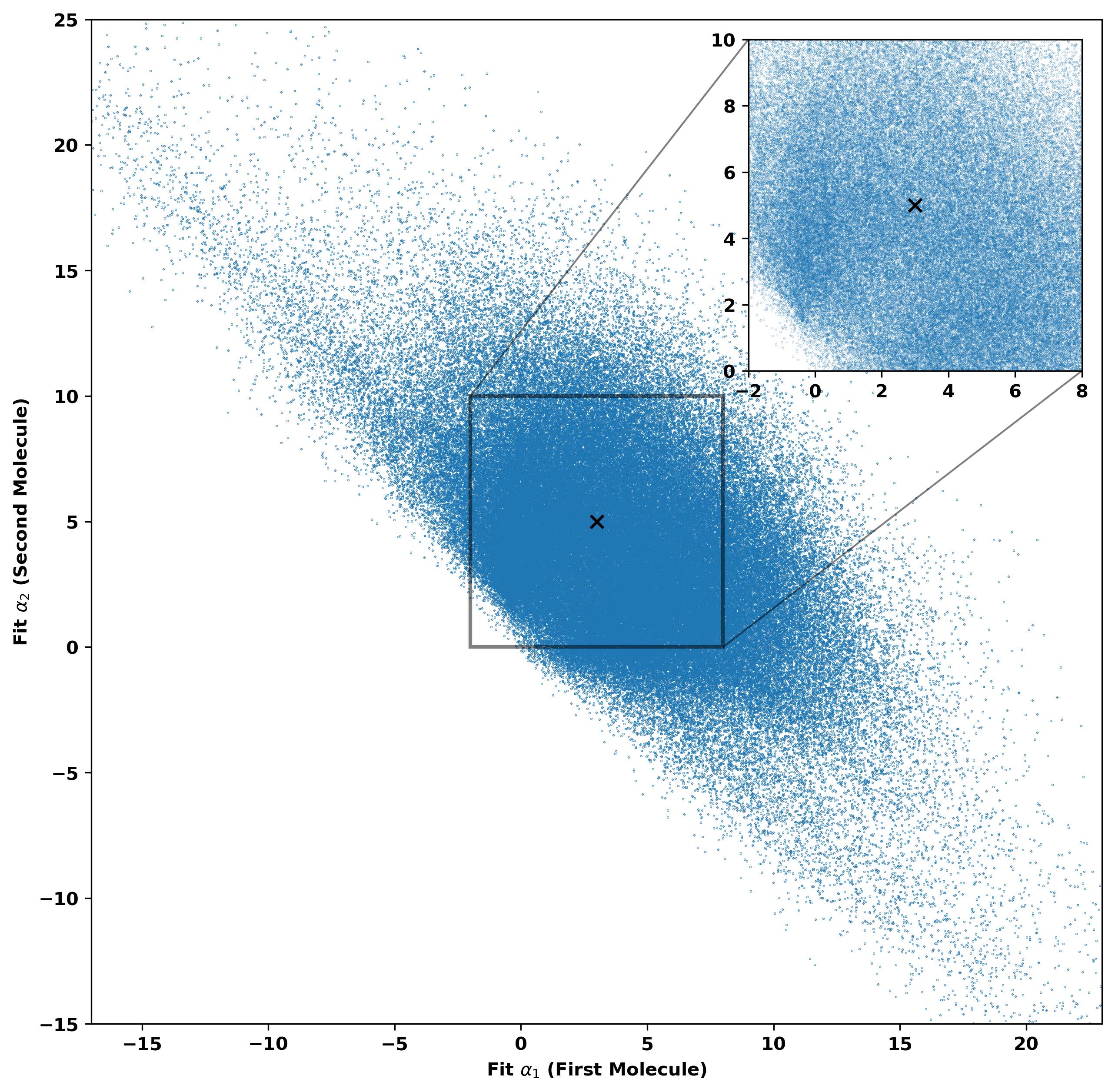}
        \caption{}
        \label{fig:comp_graph_subfig1}
    \end{subfigure}
    \hfill
    \begin{subfigure}[b]{0.45\textwidth}
        \centering
        \includegraphics[width=\textwidth]{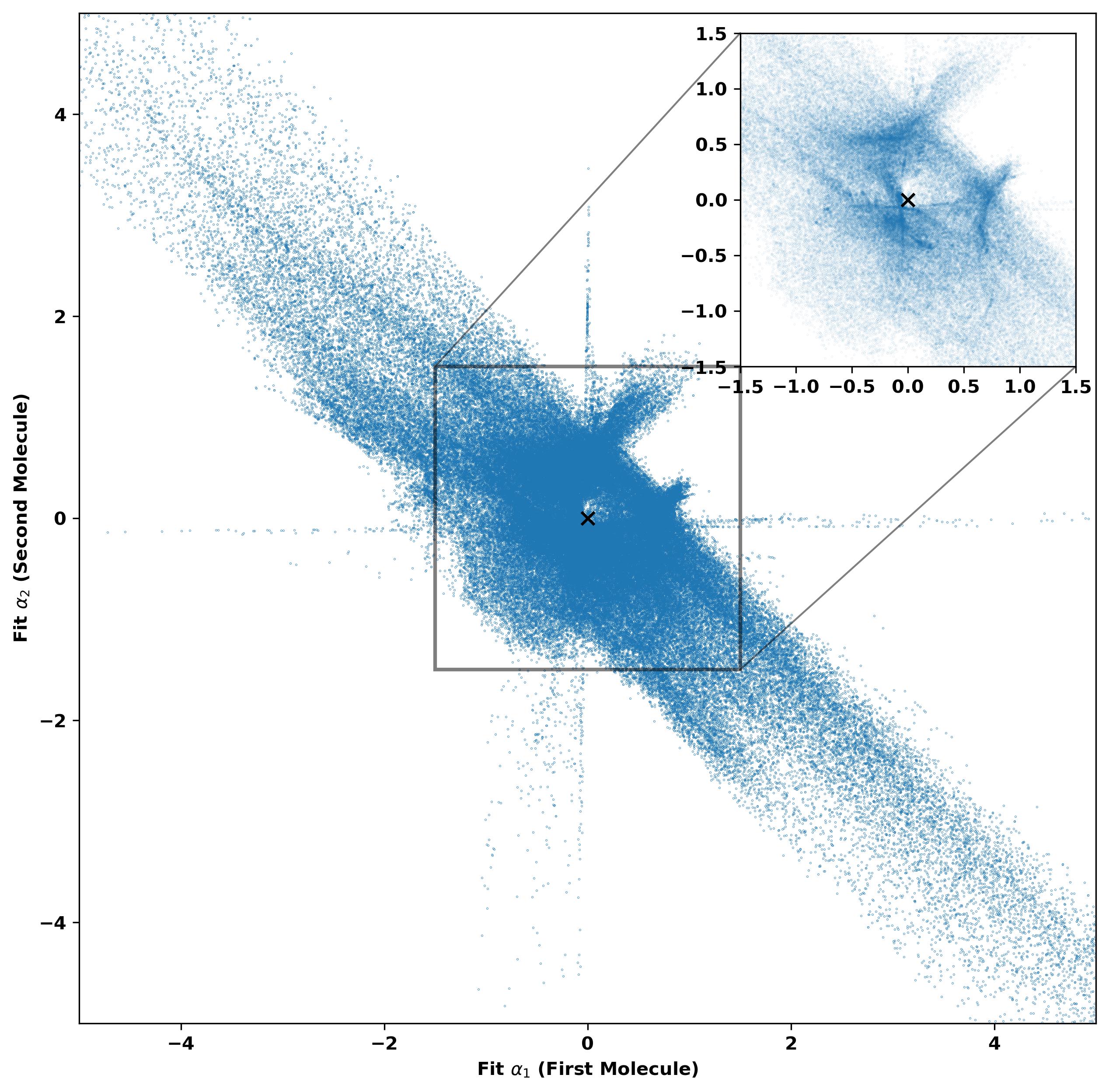}
        \caption{}
        \label{fig:regressors_fig5}
    \end{subfigure}
    \vfill
    \caption{\small\textbf{Scatter-plots of fit-coefficients for predicting the blended pair's embedding using the GNN embeddings.}\\
    \small\textbf{(a) Scatter-plot of the fit coefficient using the MPNN-GNN model from}, with zoom on centroid. Across all pairs, the average $r^2$ is $0.47$ and the $p$-value for the $F$-statistic is \SI{4.68e-5}. Notably, the distribution is not centered on the origin. In some cases, the blended pair's embedding consists of equal combinations of each individual embedding, while in other cases, one particular embedding predominates.
    \\
    \textbf{(b) Scatter-plot, as above, using the GIN-GNN embeddings.} The correlation between single molecule and blended pair embeddings was weaker, with average $r^2$ of $0.021$ and with a $p$-value for the $F$-statistic is $.445$. The distribution is centered on the origin, suggesting that for many points, neither molecule's embedding factor into the pair embedding. The vertical and horizontal lines represent where one component predominates, but the other molecule is not factored in at all.}
    \label{fig:pair_featuress}
\end{figure}

\section{Conclusion}
By applying deep-learning techniques to a novel dataset, we trained a set of models capable of accurately predicting the non-linear olfactory qualities of aroma-chemical blends. Our GNNs transfer strongly to single molecule prediction tasks. They are available on \href{https://github.com/odor-pair/odor-pair} {GitHub} for further exploration.

In our opinion, the ultimate research goal in this domain is to produce a model capable of predicting continuous labels for blends of many aroma-chemicals at varying concentrations. This mirrors the real-life work done by food scientists and perfumers.

Well-labeled public olfactory datasets that would enable this research remain scarce even for the single molecule case. Though fragrance companies likely have extensive libraries of blend recipes, these datasets remain proprietary. This paucity of public data necessitates the application of innovative techniques to boost model performance and also the exploration of novel strategies for dataset augmentation. Our work stands as a proof of concept for further research in these domains.
\subsection{Acknowledgements}
We thank Dr. Andreas Keller for mentorship and technical guidance.

\pagebreak
\setcounter{equation}{0}
\setcounter{figure}{0}
\setcounter{table}{0}
\setcounter{page}{1}
\setcounter{section}{0}
\renewcommand{\theequation}{S\arabic{equation}}
\renewcommand{\thefigure}{S\arabic{figure}}

\begin{center}
\textbf{\LARGE Supplementary Information: Olfactory}
\vspace{0.5em}
\end{center}
\begin{center}
\textbf{\LARGE Label Prediction on Aroma-Chemical }
\end{center}
\vspace{0.5em}
\begin{center}
\textbf{\LARGE Pairs}
\end{center}

\section{Methods}

\subsection{Note Canonicalization}
For labels that are difficult to predict because of their appearance across multiple structural classes, researchers and perfumers may benefit from using different labels specific to each structural class. \desc{Musk} is one such label and though \desc{floral musk} and \desc{soft musk} are frequently used to distinguish between different kinds of \desc{musky} odors, difficulty arises when \desc{musk} is used directly as a note, instead of as a family of notes. Future work could determine the feasibility of splitting this note/family apart into a number of distinct odor words. Researchers could task a panel of experts to determine if two molecules, both labelled \desc{musk}, come from the same or different structural class. If \desc{musks} from different classes are easily separable, then new descriptive words are called for.

In the same vein, our paper uses 74 labels out of a set of 109 descriptors, simply based on availability. There is no agreed on canonical set of odor descriptors, and previous works have used sets 138 labels\cite{lee2023principal}, 131 labels\cite{dravnieks1992atlas}, and as few as 19 labels\cite{keller2017predicting}. Though it is possible to predict ratings on one set of labels from another\cite{gutierrez2018predicting, sisson2022odor}, a canonical set would allow direct comparison between different approaches.
\begin{figure}[H]
    \centering
\includegraphics[width=0.5\linewidth]{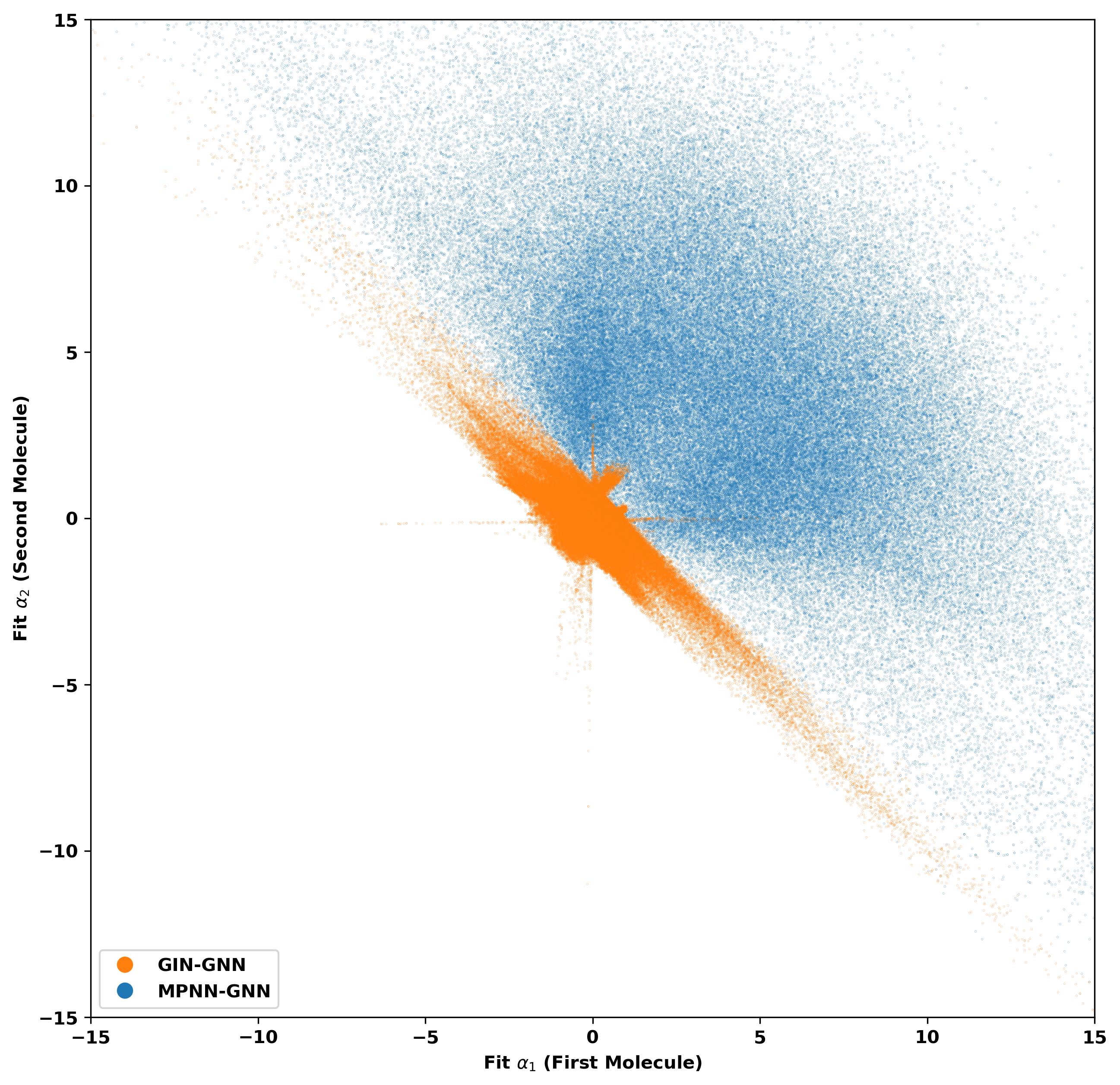}
\caption{Combined scatter-plot of the regressors for both the MPNN-GNN and the GIN-GNN.}
    \label{fig:enter-label}
\end{figure}

\begin{figure}[ht]
    \centering
    \begin{subfigure}[b]{0.45\textwidth}
        \centering
        \includegraphics[width=\textwidth]{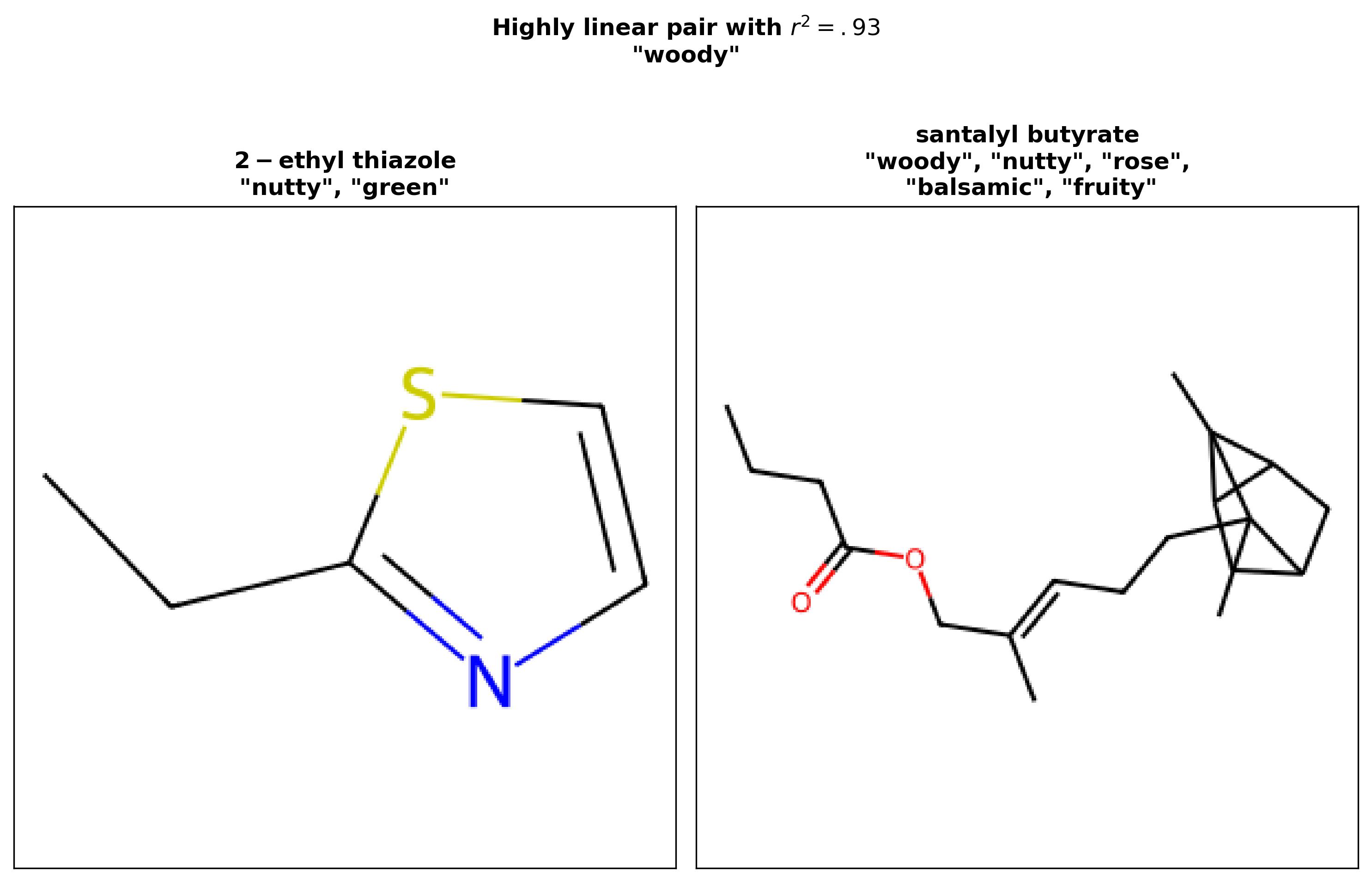}
        \caption{Highly-linear pair, where component notes simply combine.}
        \label{fig:subfig1}
    \end{subfigure}
    \hfill
    \begin{subfigure}[b]{0.45\textwidth}
        \centering
        \includegraphics[width=\textwidth]{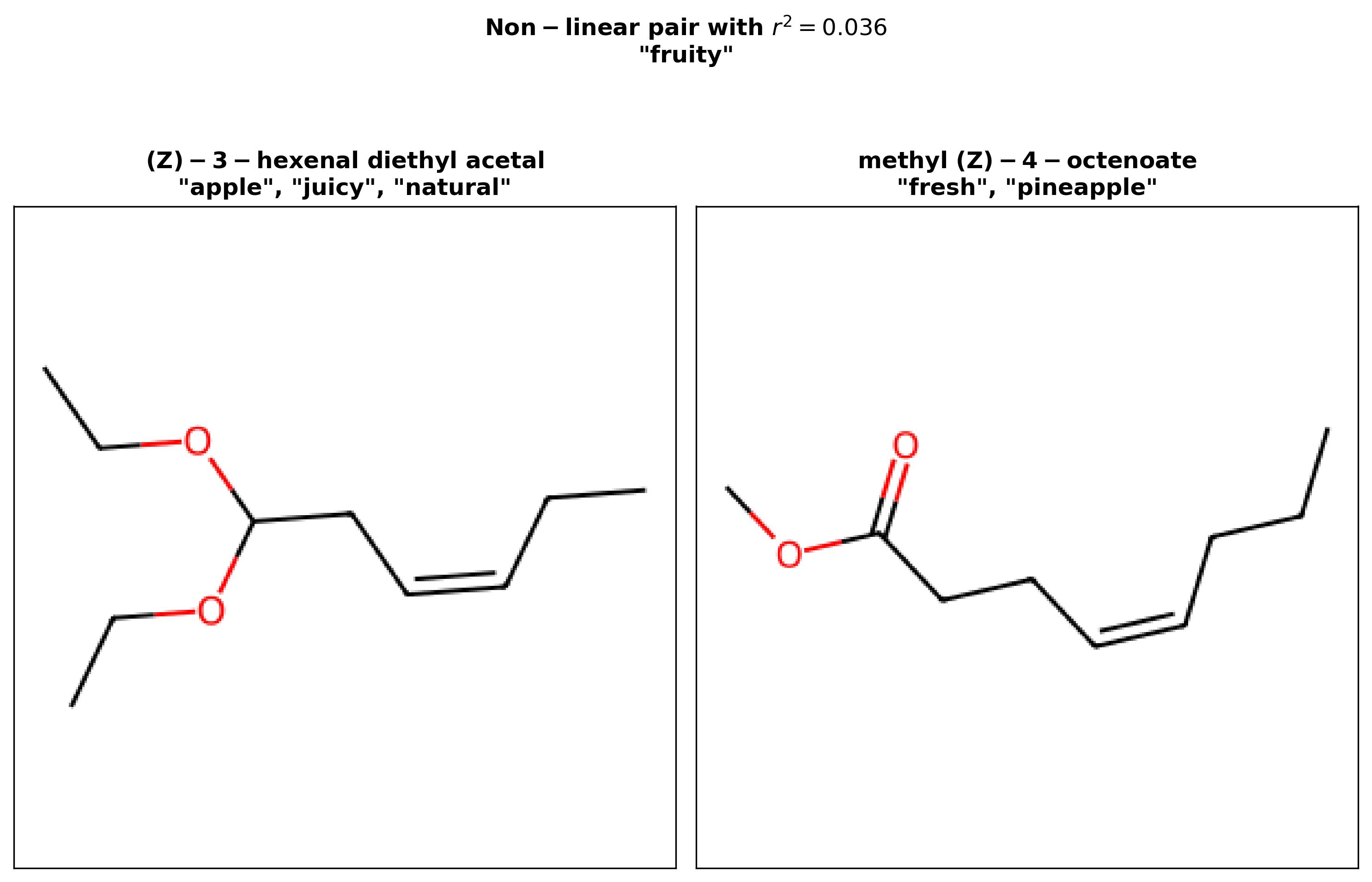}
        \caption{Non-linear pair, where emergent notes appear in the blend.}
        \label{fig:subfig2}
    \end{subfigure}
    \vfill
    \begin{subfigure}[b]{0.45\textwidth}
        \centering
        \includegraphics[width=\textwidth]{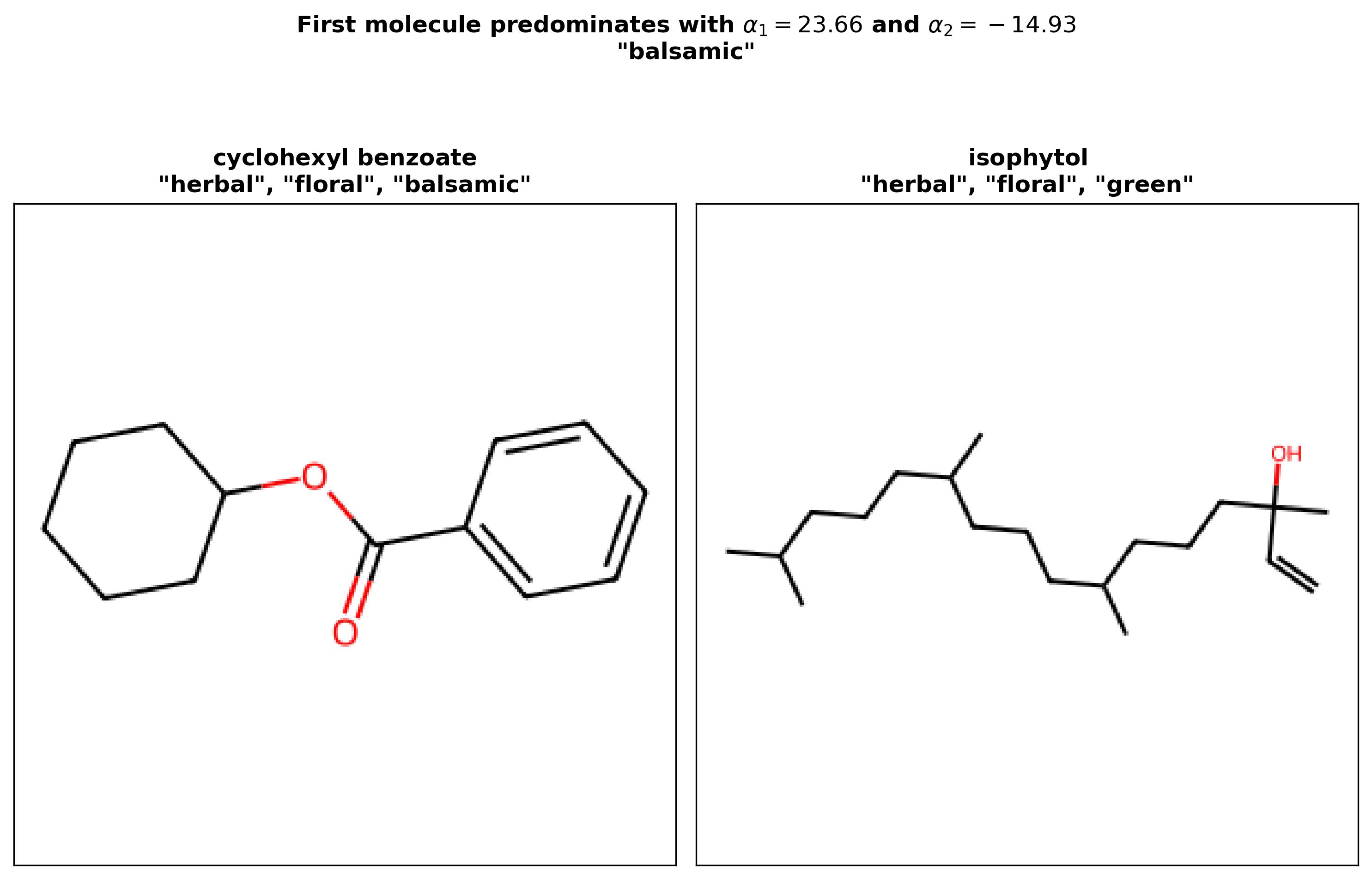}
        \caption{The first molecule predominates in this blend.}
        \label{fig:subfig3}
    \end{subfigure}
    \hfill
    \begin{subfigure}[b]{0.45\textwidth}
        \centering
        \includegraphics[width=\textwidth]{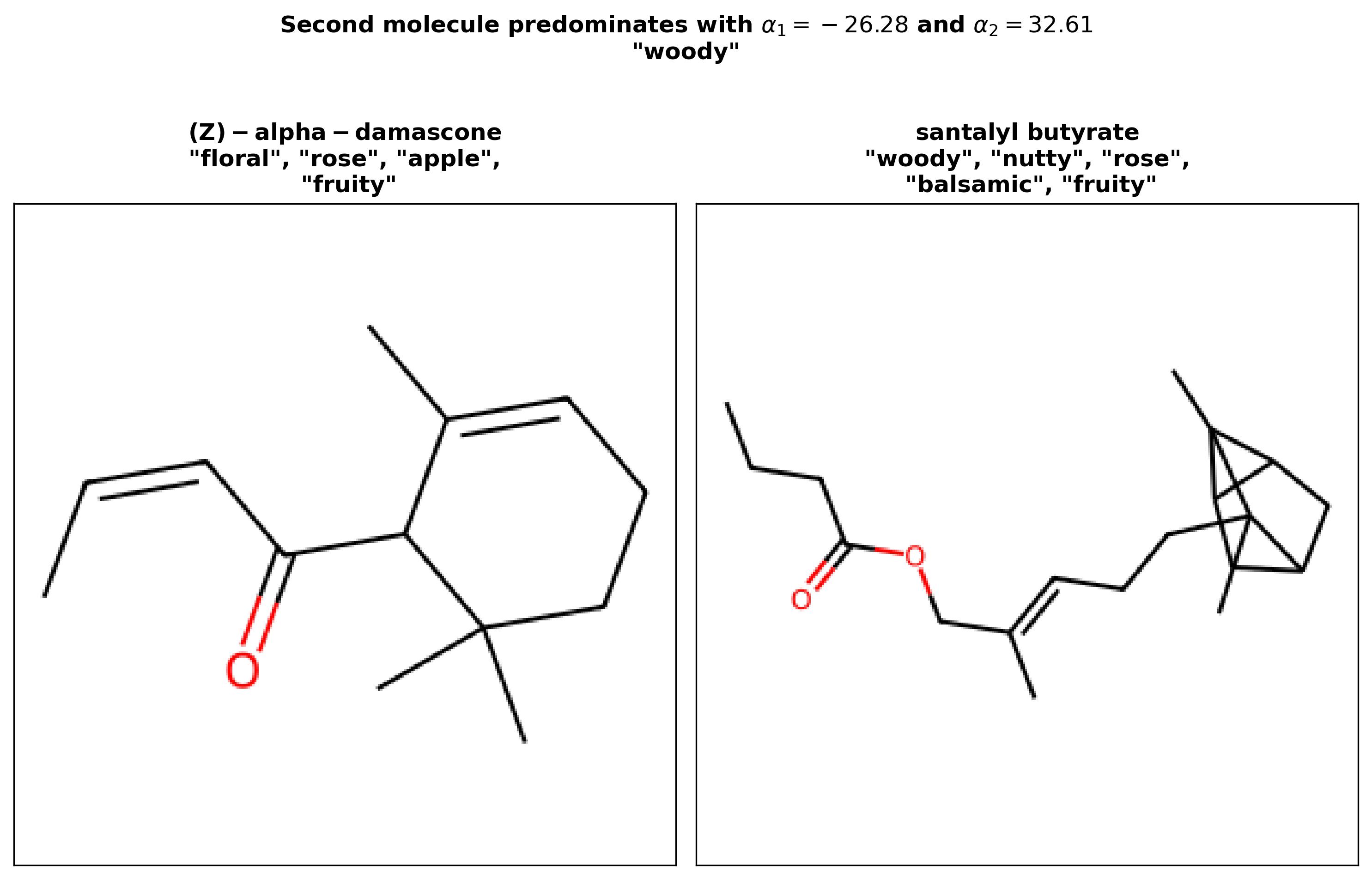}
        \caption{The second molecule predominates in this blend.}
        \label{fig:subfig4}
    \end{subfigure}
    \caption{\textbf{Four pairs of molecules, representing different blending out- comes.} While the blended results of some pairs are highly-linear, and thus the notes in the blend are easily predicted from the component molecules, other pairs are non-linear and difficult to predict. The balance between each component is also notable, as in some blends the first molecule predominates, while in others the second molecule is much more apparent.}
    \label{fig:fourplot}
\end{figure}

As shown in the Table S6 there were 11 notes that were strict subsets of other notes i.e. they exclusively linked to a parent note. While there were 3 notes as shown in Table S7 that existed independently .

The relationship between the odors of blends and their constituent molecules is intricate, as evidenced numerically by the low Jaccard coefficient observed between combinations of constituent labels and blend labels. Utilizing the set union of the labels of the constituent molecules yielded a Jaccard coefficient of 0.18, while employing the set intersection resulted in a Jaccard coefficient of 0.27. Clearly, linear prediction of blended labels from constituent labels is not feasible.

\subsection{Models and Hyperparameter Search}
Hyperparameter optimization is a challenging task, but research papers rarely provide the final hyperparameters used to train their models, and even fewer fully enumerate their search space. Much of the hyperparameter optimization procedure used here was adapted from “Neural Message Passing for Quantum Chemistry”\cite{gilmer2017neural} in which researchers built a GNN model for predicting quantum properties of organic molecules. These hyperparameters transferred well to our dissimilar task/GNN architecture, and are likely useful across disparate deep-learning chemistry domains.

\subsubsection{Graph Isomorphism Model}
\label{sec:gin}
We tested a variety of architecture using Optuna\cite{akiba2019optuna} for our hyperparameter search to find an initial, bespoke blended pair prediction model. The best architecture was structured as follows:

The Graph Isomorphism Network (GIN)\cite{xu2018powerful}, built in PyTorch Geometric\cite{fey2019fast}, was selected as the GNN architecture. The GIN was used for three message passing steps, and in order to allow weight-tying between these steps, the initial Open Graph Benchmark\cite{hu2020open} atomic encodings were padded to the hidden layer dimension ($D=832$). The carving algorithm was repeated until we generated a carving with at least one train and test data point for every label. With a 50/50 split, this took 90k iterations. Each graph carving was experimented with using Kullback–Leibler divergence, which considers the disparity between the train/test odor-label distribution and the distribution across the entire graph. However, the decision was made to prioritize optimizing for the number of usable data points rather than the similarity between components. Raising the fraction of training data meant that less data points were discarded, but also mean that it became impossible to carve a test set of molecules which covered all labels. Therefore, the 50/50 split was selected. We used a two layer feedforward neural network, built in PyTorch\cite{paszke2019pytorch}, as the update function.

The GIN generated embeddings for every atom/node in both molecules across the pair. The node-embeddings were combined using global mean and add pools concatenated together to generate graph-level embeddings, for each molecule in the pair.The graph-level embeddings for the two molecules were then further concatenated (in arbitrary order) and passed through another two-layer feedforward network to generate the pair-level embedding (also of D=832).This arbitrary ordering of graph-level embeddings worked well for the molecule-pair task, as molecules tend to appear evenly between the first and second position in the pair as a by-product of the graph-carving, but for blends of 3+ molecules, more advanced techniques, like a Set2Set\cite{vinyals2015order} model would be needed to combine the graph-level embeddings. Logits for all 74 odor labels were predicted linearly from the pair-level embeddings. We used a binary cross-entropy loss for the predicted labels with respect to their true values.
The model was scheduled for 250 epochs, but it was terminated after 121 epochs using early-stopping ($patience=0$). We used the Adam optimizer ($lr=2.1\text{e-}5$) with a decaying learning rate ($decay=0.08$) across the first 90\% of training epochs.
The entire hyperparameter search space was as follows:

\begin{table}[h]
 \captionsetup{labelformat=empty}
    \caption{Table S1: Hyperparameter values chosen as per procedure adapted in  “Neural Message Passing for Quantum Chemistry”\cite{gilmer2017neural}}
\small
\begin{tabular}{ |p{3.8cm}|p{4.5cm}|p{2.75cm}| }

\hline
 \textbf{Hyperparameter} & \textbf{Search Space} & \textbf{Chosen Value}\\
 
 \hline
 Training Epochs & [$100$,$500$] & $250$\\ \hline
 Initial Learning Rate & [$1\text{e-}4$,$1\text{e-}1$]  & $2.1\text{e-}5$\\ \hline
 Learn Rate Decay & [$1\text{e-}5$,$1\text{e-}1$]  & $2.1\text{e-}5$\\ \hline
 Hidden Dimension & [$32$,$1024$]  & $832$\\ \hline
 Feedforward Layers & [$1$,$6$]  & $2$\\ \hline
 Message Passing Steps & [$1$,$13$]  & $3$\\ \hline
 GNN Architecture & \{\textit{GIN},\textit{GCN},\textit{MPNN}\} & \textit{GIN}\\ \hline
 Aggregation  & \{\textit{Set2Set},\textit{Mean+Add Pool}\} & \textit{Mean+Add Pool}\\ 
 \hline
\end{tabular}

    \label{tab:s1}
\end{table}

\label{sec:MPNN_GNN}
\subsubsection{MPNN-GNN}
We used the MPNN-GNN model\cite{lee2023principal}, which was originally trained on 138 odor descriptors using LF-GS combined dataset [2] to predict odors of single molecules. The model contains message passing layers, a readout layer with radius 0 combination to fold atom and bond embeddings together followed by set2set operation and a feed-forward network layer with sigmoid activation for multi-label classification predictions.

A summed binary cross-entropy loss is used in the training process for every descriptor. Each descriptor's influence is taken into account in the loss computation, and its weight is determined by a factor of log(1+Imbalance ratio per label). For every label $\lambda$ in a multi-label dataset M, the Imbalance ratio per label (IRLbl) is calculated, where $Y_i$ is the label-set of the $i^{th}$ instance. The ratio of the specific label $\lambda$ to the majority label is used to calculate IRLbl. For the most common label, it takes on a value of 1, while for other labels, it takes on a greater value. The imbalance for the associated label increases with the IRLbl value.

The hyperparameter search space is given in Table S2.

\begin{table}[h]
 \captionsetup{labelformat=empty}
    \caption{Table S2: Best hyperparameter values chosen  from experiments on the LF-GS dataset for mixture model odor prediction}

\small
\begin{tabular}{ |p{4.2cm}|p{4.9cm}|p{2.65cm}| }
\hline
 \textbf{Hyperparameter} & \textbf{Search Space} & \textbf{Chosen Value}\\
 
 \hline
ffn Dropout probability & [$0.12$,$0.25$,$0.5$] & $0.12$\\ \hline
Weight decay & [$1\text{e-}3$,$1\text{e-}4$, $1\text{e-}5$ ]  & $1\text{e-}5$\\ \hline
Initial learning rate & [$0.001$,$0.0001$, $0.005$,$0.0005$]  & $0.001$\\ \hline
 Decay Rate & [$0.25$,$0.5$,$0.75$]  & $0.5$\\ \hline
 Decay Steps & [$42*5$,$42*10$, $42*15$, $42*20$]  & $42*20$\\ \hline
 Number of step set2set  & [$2$,$3$, $4$]  & $3$\\ \hline
 Number of layers set2set & [$2$,$3$,$4$], & $3$\\ \hline
 FFN hidden list & [[$200$], [$60$, $60$], [$300$], [$500$, $500$], [$300$, $300$]] & $[$300$]$\\ \hline

 Aggregation  & \{\textit{global sum pooling},\textit{set2set}\} & \textit{set2set}\\
 \hline
\end{tabular}

\label{mpnn_gnn}
\end{table}

\subsubsection{Single Molecule Trained Model}
In this study, we aim to forecast the odors of molecular mixtures. Our dataset of odorant mixtures includes 74 odor notes, with 60 of these notes also present in the LF-GS combined dataset.In the experiment with the model trained on single molecules, we utilize the LF-GS dataset for training and test it on a 50:50 split of the mixture dataset. Given that the odorant mixture dataset was derived from Goodscents data, part of the LF-GS combined dataset, we remove common SMILES from the test set in the LF-GS dataset. Consequently, 60 out of 138 odor notes are used. This subset of the LF-GS dataset is then divided into training and validation sets in an 80:20 ratio using random stratified splitting.
\begin{table}[h]
 \captionsetup{labelformat=empty}
    \caption{Table S3: Best hyperparameter values chosen  from experiments on the LF-GS dataset for single molecule trained model prediction}

\small
\begin{tabular}{ |p{4.2cm}|p{4.9cm}|p{2.60cm}| }
 \hline
 \textbf{Hyperparameter} & \textbf{Search Space} & \textbf{Chosen Value}\\
 
 \hline
ffn Dropout probability & [$0.12$,$0.25$,$0.5$] & $0.25$\\\hline
Weight decay & [$1\text{e-}3$,$1\text{e-}4$, $1\text{e-}5$ ]  & $1\text{e-}5$\\\hline
Initial learning rate & [$0.001$,$0.0001$, $0.005$,$0.0005$]  & $0.005$\\\hline
 Decay Rate & [$0.25$,$0.5$,$0.75$]  & $0.75$\\\hline
 Decay Steps & [$4*5$,$4*10$, $4*15$, $4*20$]  & $4*10$\\ \hline
 Number of step set2set  & [$2$,$3$, $4$]  & $2$\\ \hline
 Number of layers set2set & [$2$,$3$,$4$], & $2$\\ \hline
 FFN hidden list & [[$200$], [$60$, $60$], [$300$], [$500$, $500$], [$392$, $392$]] & $[$200$]$\\ \hline

 Aggregation  & \{\textit{global sum pooling},\textit{set2set}\} & \textit{set2set}\\
 \hline
\end{tabular}
\label{mpnn_single}
\end{table}
We proceeded with using the best parameters from experiments on the LF-GS dataset for single molecule odor prediction.
(Note: The model hyperparameters could have been fine tuned on 100 trials of random search)3 models with different random seeds were trained to predict odors of single molecules. Then the models were tested on Test split from odor mixture dataset to assess how the model performs on predicting odors of mixture of molecules.

We observed that the mean ROC score for Test split (average of 3 models) is 0.8029 ±0.00147 (95\% confidence level).
The ensemble of these 3 models gave ROC score for Test set = 0.8146.We have 5 folds of train-valid-test split of the odor mixture set with odor notes in the range of 41 to 44 per fold. The assumption is that the best hyperparameters from that run would show similar results with 50:50 split with 74 odors. The reason we had to do this was the difficulty in extracting 5 fold train-valid-test splits with 74 odors notes, due to the limitation of the graph carving and nature of the mixture datasets.We performed 100 trials of random search cross validation to get the best values for hyperparameters like regularization weight decay, set2set readout parameters, dropout and hidden layer size of feed forward layer, initial learning rate, decay rate and decay step for exponential decay learning rate scheduling. The best validation ROC score is 0.7928 and best test ROC score is 0.7806.

\textbf{Note:} We used the best hyperparameters for message passing layers for MPNN-GNN from single molecule training results.
\section{Supplementary Data}
\begin{table}[!ht]
 \captionsetup{labelformat=empty}
    \caption{Table S4: Top 10 predicted labels by the model}

    \centering
    \begin{tabular}{ |p{2cm}|p{2.6cm}|p{2.4cm}| p{3.6cm}| }
    \hline
      \textbf{Label} & \textbf{MPNN-GNN}  & \textbf{MPNN-GIN } & \textbf{Morgan Fingerprint} \\ \hline
        acidic & 0.98 & 0.99 & 0.94 \\ \hline
        cheesy & 0.95 & 0.97 & 0.98 \\ \hline
        mossy & 0.94 & 0.97 & 0.94 \\ \hline
        sour & 0.99 & 1.00 & 0.80 \\ \hline
        vanilla & 0.98 & 0.92 & 0.97 \\ \hline
        amber & 0.95 & 0.96 & 0.86 \\ \hline
        coffee & 0.94 & 0.92 & 0.88 \\ \hline
        buttery & 0.95 & 0.93 & 0.83 \\ \hline
        fresh & 0.91 & 0.90 & 0.92 \\ \hline
        chocolate & 0.95 & 0.81 & 0.97 \\ \hline

    \end{tabular}
\end{table}
\begin{table}[!ht]
 \captionsetup{labelformat=empty}
    \caption{Table S5: Bottom 10 predicted labels by the model}

    \centering
    \begin{tabular}{ |p{2cm}|p{2.6cm}|p{2.4cm}| p{3.6cm}| }
    \hline
      \textbf{Label} & \textbf{MPNN-GNN}  & \textbf{MPNN-GIN } & \textbf{Morgan Fingerprint} \\ \hline
        dairy & 0.56 & 0.43 & 0.12 \\ \hline
        earthy & 0.37 & 0.42 & 0.44 \\ \hline
        orris & 0.29 & 0.46 & 0.45 \\ \hline
        cocoa & 0.57 & 0.56 & 0.29 \\ \hline
        rummy & 0.56 & 0.45 & 0.47 \\ \hline
        tropical & 0.52 & 0.55 & 0.45 \\ \hline
        bitter & 0.45 & 0.17 & 0.58 \\ \hline
        cooling & 0.35 & 0.33 & 0.58 \\ \hline
        musty & 0.62 & 0.57 & 0.48 \\ \hline
        sweet & 0.58 & 0.17 & 0.65 \\ \hline

    \end{tabular}
\end{table}

\begin{table}[ht]
 \captionsetup{labelformat=empty}
    \caption{Table S6: Notes that occurred exclusively in association with a parent note}

    \centering
    \begin{tabular}{ |p{2.89cm}|p{2.15cm}|p{1.65cm}| p{1.1cm}| }
    \hline
       \textbf{Dependent Note}  & \textbf{Parent Note } & \textbf{Frequency } \\ \hline
        estery & fruity & 739\\ \hline
        cherry & floral & 443  \\ \hline
        toasted & butterfly & 218  \\ \hline
        juicy & sulfurous & 185  \\ \hline
        tomato & vegetable & 138  \\ \hline
        tobacco & nutty & 101  \\ \hline
        potato & vegetable & 51  \\ \hline
        celery & lactonic & 34  \\ \hline
        lactonic & celery & 34  \\ \hline
        dusty & fruity & 32  \\ \hline
        tarragon & anise & 28  \\ \hline
    \end{tabular}
    \label{table:notes_name}
\end{table}
\begin{table}[!ht]
\captionsetup{labelformat=empty}
    \caption{Table S7: Isolate Note with the frequency}

    \centering
    \begin{tabular}{ |p{2.15cm}|p{1.65cm}|p{4.5cm}| }
    \hline
       \textbf{Isolate Note}  & \textbf{Frequency}  \\ \hline
        ammoniacal & 9 \\ \hline
        salty & 1  \\ \hline
        hay & 2 \\ \hline
        
    \end{tabular}
    \label{table:freq_notes_name}
\end{table}

\section{Results}
\subsection{MPNN-GNN Performance}

To understand the performance of the MPNN model at the level of individual labels, we carried out a series of experiments. In this study, we investigate the targets of the worst and best performing odor labels, as well as the actual predictions made by the model. We aim to analyse how model predictions correlate with odor descriptors of individual components within the mixture, both in the training and test sets.

 The inference that we could draw from the experiments was that the top and bottom 5 performances are not correlated to training data size. While Lee et al. \cite{lee2023principal} observed a notable improvement in model performance with an increased number of training examples, our study did not replicate this finding. The insights derived from analyzing the results of the odor mix model for the bottom five predicted labels are as follows: \textbf{a) Orris:} Our analysis revealed that 20\% of \desc{orris} samples in the test set also have \desc{spicy} as a label. Despite significant representation of \desc{orris} in both the training and test sets, there's a notable imbalance: 42 \desc{orris} samples in the training set compared to 1238 \desc{spicy} samples. This imbalance could potentially cause the model to favour \desc{spicy} as the primary descriptor for mixtures containing \desc{orris}. \textbf{b) Cooling:} In the test set, approximately 75\% of the \desc{cooling} target distribution is \desc{green} and less than 20\% is \desc{fruity}. However, the model predicts about 50\% as \desc{estery} and 30\% as \desc{fruity}. Despite \desc{green} being a major descriptor in both the training and test sets, predictions for \desc{green} are less than 2\% for mixtures with a \desc{cooling} target. The model appears biased towards component odors not present in the training set mixtures. \textbf{c) Earthy:} For the \desc{earthy} target distribution, around 20\% are \desc{musty} and \desc{waxy} in the test set. However, predictions vary: 40\% are predicted as \desc{fatty}, 30\% as chocolate, and only a small share as \desc{waxy} or musty. \desc{waxy} is well-predicted for mixtures that smell both \desc{earthy} and \desc{waxy}. Both the training and test sets have high \desc{earthy} content, indicating no bias. \textbf{d) Melon:} For the \desc{melon} target distribution, there is a very small percentage of \desc{floral} and \desc{fruity} in the test set. Model predictions are poor, with over 40\% predicted as \desc{creamy}. Both the training and test sets have high \desc{melon} content, indicating no bias. \textbf{e) Bitter: }For the \desc{bitter} target distribution, there is a small percentage of \desc{waxy} and \desc{fatty} in the test set. The model most often predicts cheesy and sour. The test set has higher cheesy content, with no components smelling \desc{bitter}, indicating that \desc{bitter} might be an emergent odor. The model seems biased towards component odors not in the training set mixtures. Less than 50\% of the training set mixtures have a \desc{bitter} odor in any component.

From the analysis of the odor mix model, the top five predicted labels provide the following insights: \desc{Sour}, \desc{acidic}, \desc{vanilla}, \desc{mossy}, and \desc{cheesy} perform as expected, with the highest prediction scores for their respective odor labels. Even their mixture components exhibit a high percentage of these odor labels.

The inferences drawn from the analysis of the odor single model results on the mixture test set for the five bottom predicted labels are: \textbf{a) Bitter:} For the \desc{bitter} target distribution, the test set has a small percentage of \desc{waxy} and \desc{fatty}. However, the model predicts \desc{sour}, \desc{winey}, and \desc{buttery} with the highest score. None of the mixture components in the test set have a \desc{bitter} smell, explaining the poor performance of \desc{bitter} with a ROC score of \(< 0.1\). The model seems to predict mixture odors based on individual components. \textbf{b) Melon:} For the \desc{melon} target distribution, the test set has minimal \desc{floral} and \desc{fruity} percentages. Model predictions for \desc{melon} mixtures are quite poor, with little more than 50\% predicted as coumarinic. The training (single molecules) and test mixture components have high \desc{melon} content. \textbf{c) Berry:} For the \desc{berry} target distribution, the test set has a small percentage of \desc{floral} and \desc{fruity}. Model predictions for \desc{berry} mixtures are quite poor, with over 35\% of predictions not assigning any odor label. Interestingly, the training (single molecules) and test mixture components have high \desc{berry} content. \textbf{d) Oily:} For the \desc{oily} target distribution, the test set has minimal \desc{fruity}, \desc{ethereal}, \desc{green}, and rummy odors. Model predictions are poor, with over 25\% not assigning any odor label. Interestingly, mixtures with \desc{oily} odors containing \desc{rummy} and \desc{ethereal} are still predicted accurately. The training and test mixture components have high \desc{oily} content.
\textbf{e) Orris:} For the \desc{orris} target distribution, 20\% also have \desc{spicy} in the test set. Model predictions for \desc{orris} mixtures are poor, with over 40\% not assigning any odor label and around 40\% predicted as \desc{spicy}. Interestingly, the training and test mixture components have high \desc{orris} content.

The top 5 predicted labels from the odor mix model analysis reveal that \desc{sour}, \desc{onion}, \desc{meaty}, \desc{alliaceous}, and \desc{roasted} perform as expected, with the highest prediction scores for their respective odor labels. Even their mixture components show a high percentage of these odor labels. However, it's worth noting that odors like \desc{onion}, \desc{alliaceous}, and \desc{roasted} have a second set of odor labels with high co-occurrence in the single molecules dataset. While this strong co-occurrence isn't observed in mixture labels, the model predicts them with high scores. This suggests the model may struggle to prioritize dominant odors in mixtures, often predicting labels of individual components.

In our analysis, we aimed to address two questions: Firstly, how GNNs capture the relationship between molecular structure and odor; secondly, how GNNs combine their underlying models, captured in the embedding space, for individual molecules into blended pairs. To answer the second question, we analyzed the embedding space.
\bibliographystyle{unsrt}
\bibliography{paperpile}

\end{document}